\documentclass[fleqn,10pt]{wlscirep}
\usepackage[utf8]{inputenc}
\usepackage[T1]{fontenc}

\usepackage{multirow}
\usepackage{amsmath,amssymb,amsfonts}
\usepackage{amsthm}
\usepackage{mathrsfs}
\usepackage{textcomp}
\usepackage{booktabs}
\usepackage{tabularx}
\usepackage{array}
\newcolumntype{L}[1]{>{\raggedright\arraybackslash}p{#1}}
\newcolumntype{Y}{>{\raggedright\arraybackslash}X}
\usepackage{ragged2e}
\usepackage{placeins}
\usepackage{float}
\usepackage[font=small,labelfont=bf,skip=2pt]{subcaption}

\newtheorem{theorem}{Theorem}
\newtheorem{proposition}[theorem]{Proposition}
\newtheorem{lemma}[theorem]{Lemma}
\newtheorem{definition}[theorem]{Definition}
\newtheorem{remark}[theorem]{Remark}

\title{Inner-Speech Decoding from EEG Across Three Corpora Under a Fixed Architecture and Protocol}

\author[1,*]{Md Wahiduzzaman Suva}
\author[1]{Md. Imtiaj Alam Sajin}
\author[1]{Esm E Moula Chowdhury Abha}
\author[1]{Susham Moula Choudhury Akash}
\author[2]{Md Tauhidul Islam}
\author[1]{Abhijit Bhowmik}
\author[1]{Md Manirul Islam}
\author[1,3]{Debajyoti Karmaker}

\affil[1]{Computer Science Department, American International University-Bangladesh, Dhaka, Bangladesh}
\affil[2]{Department of Software Engineering, N.I. Lobachevsky National Research Nizhny Novgorod State University, Nizhny Novgorod, Russia}
\affil[3]{School of Computing, RMIT University, Melbourne, Australia}

\affil[*]{d.karmaker@aiub.edu}


\begin{abstract}
Inner-speech decoding from electroencephalography could support non-invasive communication for people with severe motor impairment, but comparison across studies is hindered by dataset- and protocol-specific evaluation. We present a controlled cross-dataset study using one preprocessing pipeline, one within-subject five-fold cross-validation protocol, and a compact dual-attention Transformer (VC-DATNet; approximately 131,000 parameters) across three public corpora: Thinking Out Loud (10 participants), BISpeech (4), and SyncSpeech (3). VC-DATNet achieved per-participant sensor/source hybrid accuracies of 39.48\% on the four-class Thinking Out Loud task, 20.60\% on the eight-word BISpeech task, and 25.53\% on the eight-word SyncSpeech task, for which we report the first decoding result. Sensor-space and cortical source-space representations were complementary, with each performing better for different participants; this pattern replicated across all three corpora and nine architectures. Moreover, ten decoders spanning convolutional, recurrent, temporal-convolutional, hybrid Transformer, and pretrained foundation-model families converged within a narrow accuracy range on every dataset. These findings indicate that representation choice should be individualized and suggest that performance on these public corpora is constrained more by signal quality and dataset scale than by model capacity. The released pipeline provides a reproducible benchmark for future inner-speech decoding studies.
\end{abstract}

\begin{document}

\raggedbottom
\maketitle
\thispagestyle{empty}

\section*{Introduction}\phantomsection \label{sec:introduction}

Electroencephalography (EEG) provides millisecond-resolution recordings of cortical activity through non-invasive scalp electrodes and underpins a growing class of brain--computer interface (BCI) systems aimed at restoring communication for individuals with severe motor impairment~\cite{lotte2018review,roy2019deep,vafaei2025transformers,imaginedspeech_review2025}. Within this space, inner speech decoding (the classification of silent, internally articulated verbal commands directly from EEG) offers a more naturalistic and cognitively expressive interaction channel than the established motor imagery (MI) paradigm. Rather than mapping abstract limb movements to symbolic control signals, an inner speech BCI translates the user's intended words. The relevance is direct in late-stage amyotrophic lateral sclerosis, locked-in syndrome, and post-stroke aphasia, where overt motor pathways are unavailable~\cite{nieto2022thinking}.

Decoding inner speech from EEG is qualitatively harder than decoding motor imagery, and the gap is structural rather than incidental. First, inner speech engages distributed and partially overlapping cortical networks (inferior frontal, superior temporal, parietal), so the discriminative signal is spatially diffuse rather than focal as in sensorimotor mu/beta desynchronisation~\cite{nieto2022thinking}. Second, the underlying neural amplitudes are markedly lower, pushing signal-to-noise ratios towards the limits of scalp EEG resolution. Third, within-subject trial-to-trial variability is higher because subjects cannot rehearse identical motor outputs~\cite{Gasparini2022InnerSpeech}. End-to-end inner speech accuracies on public datasets therefore remain in the 30--45\,\% range for four-class problems, well below the 80\,\%+ regime now routine for motor imagery~\cite{nieto2022thinking,Gasparini2022InnerSpeech,imaginedspeech_review2025}.

Architectural progress in deep-learning-based EEG decoding has nevertheless been driven almost exclusively by motor-imagery benchmarks: compact convolutional models~\cite{lawhern2018eegnet,schirrmeister2017deep,ingolfsson2020eeg}, multi-scale CNNs~\cite{riyad-2020,tang2023motor,ma2024attention,xie2024bfatcnet,wu2025ameegnet,liao2025ciacnet}, hybrid CNN-Transformers~\cite{zhao2024ctnet,ding-2024,zhao2025mscformer,altaheri2025tcformer,liao2025eegencoder,satransnet2025,cltnet2025}, and dual-attention frameworks~\cite{du2022dual,pan-2024,liu2025dmsacnn,li2024dual,han2024edpnet} are developed and ranked almost exclusively on BCI Competition IV-2a/2b. Inner speech has received comparatively little attention, and the studies that do exist are fragmented: each is typically reported on a single corpus, under a single evaluation protocol, with a single decoder, which makes it difficult to separate transferable findings from dataset- or model-specific artefacts~\cite{nieto2022thinking,Gasparini2022InnerSpeech,imaginedspeech_review2025}. Three gaps follow. First, no published deep architecture has been evaluated on more than one inner-speech corpus, so within-paradigm replication, the standard form of external validity, is absent. Second, the choice of signal representation is rarely studied: most decoders operate on sensor-space EEG, yet volume conduction makes a cortical source-space projection a plausible alternative, and the two have not been compared systematically on this paradigm. Third, it is not established whether the modest accuracies reported for inner speech reflect a limitation of the decoders used or a ceiling set by the signal itself.

This paper addresses these gaps with a deliberately controlled study. We fix one compact decoder and one preprocessing and evaluation protocol, and apply them unchanged across three independent public inner-speech EEG datasets. Holding the architecture constant is the central methodological choice: it ensures that any difference observed between datasets, between sensor- and source-space representations, or between subjects is attributable to the data and the representation rather than to model selection. The decoder, VC-DATNet, is a compact dual-attention Transformer (a multi-scale temporal-convolution front end, a dual mean-and-log-variance pooling stage, and a gated self-attention encoder) in the now-common family of transformer-based EEG decoders~\cite{zhao2024ctnet,han2024edpnet}; its variance-pooling stage is motivated by the fact that event-related (de)synchronisation manifests as a modulation of band-limited power rather than of mean amplitude, a connection formalised in \hyperref[subsec:dual_pooling]{Dual Pooling Module}. Its role in this study is as a consistent, reproducible instrument rather than as a performance claim.

The three datasets are the Thinking Out Loud corpus (ToL; OpenNeuro ds003626)~\cite{nieto2022thinking}, the EEG modality of the Bimodal Inner Speech corpus (BISpeech; OpenNeuro ds004196)~\cite{liwicki2023openneuro,simistira2023bimodal}, and the Synchronous EEG--fMRI Inner Speech corpus (SyncSpeech; OpenNeuro ds006033)~\cite{simistira2025ds006033}, decoded here for the first time. They differ in cohort size (10, 4, and 3 subjects), stimulus type (directional commands vs.\ semantic words), channel count, and acquisition hardware, so behaviour that is consistent across all three is evidence of a genuine paradigm-level effect rather than a single-dataset artefact. The principal contributions are:

The principal contributions of this work are as follows:

\begin{enumerate}
    \item We conduct the first controlled cross-dataset evaluation of inner-speech EEG decoding, applying a single architecture and a single preprocessing and evaluation pipeline to three independent public corpora (ToL~\cite{nieto2022thinking}, BISpeech~\cite{liwicki2023openneuro,simistira2023bimodal}, and SyncSpeech~\cite{simistira2025ds006033}) under within-subject five-fold cross-validation. To our knowledge this is also the first decoding result reported on SyncSpeech (\hyperref[subsec:innerspeech_results]{Inner-Speech Decoding Across Three Datasets}).

    \item We identify a two-regime sensor/source effect: sensor-space decoding and decoding from a cortical source-space projection are complementary, each yielding the higher accuracy on a distinct subset of subjects, with neither representation dominating. The effect appears independently on all three datasets, establishing it as a property of inner-speech EEG rather than of any single corpus (\hyperref[subsec:innerspeech_results]{Inner-Speech Decoding Across Three Datasets}).

    \item We provide evidence for a signal-limited accuracy ceiling. With the architecture and protocol held fixed, ten decoders spanning five method families (convolutional, temporal-convolutional, recurrent, hybrid CNN--Transformer, and a 25-million-parameter pretrained foundation model) converge to the same narrow accuracy band on every dataset, consistent with a signal-limited rather than capacity-limited ceiling on these public corpora (\hyperref[subsec:ceiling]{A Signal-Limited Accuracy Ceiling}).

    \item We benchmark VC-DATNet against the strongest peer-reviewed baseline on each dataset and, for ToL, independently reproduce the strongest published baseline (Gasparini et al.~\cite{Gasparini2022InnerSpeech}) under our own pipeline to within 0.1 percentage points of the reported value, verifying that every headline comparison is fair (\hyperref[subsec:innerspeech_results]{Inner-Speech Decoding Across Three Datasets}).

    \item We introduce VC-DATNet, a compact ($\approx$131K-parameter) dual-attention Transformer whose design choices are formally grounded by Lemma~\ref{lem:logvar_psd} (log-variance pooling estimates band-limited log-power) and Proposition~\ref{prop:simplex} (pre-softmax value-gating preserves the attention probability simplex). The complete preprocessing pipeline, source-space projection code, model implementation, and per-run training logs are released publicly (\hyperref[subsec:impl]{Implementation and Hyperparameters}).
\end{enumerate}

\section*{Related Work}\phantomsection \label{sec:related}

We survey the work most relevant to VC-DATNet in four families (CNN-based decoders, Transformer and hybrid CNN--Transformer models, dual-attention and variance-sensitive feature extraction, and inner-speech decoding), each subsection closing with an explicit ``gap that motivates VC-DATNet'' paragraph. Tables~\ref{tab:related_eeg} and~\ref{tab:related_innerspeech} give a compact comparative summary of the studies discussed, including the recent works most directly comparable to ours, so that the positioning of VC-DATNet against the current literature can be read at a glance.

\begin{table}[!htbp]
\centering
\scriptsize
\setlength{\tabcolsep}{3pt}
\renewcommand{\arraystretch}{1.08}
\caption{Comparative summary of representative EEG decoding architectures (2017--2025) surveyed in \hyperref[sec:related]{Related Work}, included as a literature reference for the EEG decoder families discussed below. Accuracies are not reported in this table because nearly every architecture listed was developed and evaluated on motor-imagery benchmarks; retraining them under our pipeline on the three inner-speech corpora used in this paper is a direct extension (\hyperref[subsec:limitations]{Limitations and Scope}). Inner-speech / imagined-speech studies are summarised separately in Table~\ref{tab:related_innerspeech}. Architectures are grouped by family. MI = motor imagery; HGD = High-Gamma Dataset.}
\label{tab:related_eeg}
\begin{tabularx}{\textwidth}{@{}L{2.4cm} Y L{3.0cm} Y@{}}
\toprule
\textbf{Study} & \textbf{Method(s)} & \textbf{Original evaluation} & \textbf{Highlights / relevance} \\
\midrule
\multicolumn{4}{@{}l}{\textit{CNN-based decoders}} \\
\midrule
Schirrmeister et al.~\cite{schirrmeister2017deep} & ShallowConvNet / DeepConvNet & BCI~IV-2a, HGD & End-to-end CNNs match filter-bank pipelines \\
Lawhern et al.~\cite{lawhern2018eegnet} & EEGNet (depthwise-separable conv) & BCI~IV-2a, others & Compact, cross-paradigm default baseline \\
Ingolfsson et al.~\cite{ingolfsson2020eeg} & EEG-TCNet (temporal conv blocks) & BCI~IV-2a & Low-latency convolutional decoder \\
Riyad et al.~\cite{riyad-2020} & Incep-EEGNet (inception filters) & BCI~IV-2a & Multi-scale temporal receptive fields \\
Mane et al.~\cite{mane2021fbcnet} & FBCNet (filter-bank + variance layer) & BCI~IV-2a/2b & Explicit spectral-variance layer \\
Tang et al.~\cite{tang2023motor} & Multi-scale conv + frequency enhancement & BCI~IV & Multi-scale + frequency-domain fusion \\
Ma et al.~\cite{ma2024attention} & Attention-augmented CNN (TransNet) & MI & Multi-modal temporal fusion \\
Xie et al.~\cite{xie2024bfatcnet} & BFATCNet (bidirectional feature pyramid + TCN) & MI & Bottom-up/top-down band propagation \\
Wu et al.~\cite{wu2025ameegnet} & AMEEGNet (3 EEGNet branches + ECA) & BCI~IV-2a/2b, HGD & Parallel multi-scale + channel attention \\
Liao et al.~\cite{liao2025ciacnet} & CIACNet (dual-branch CNN + CBAM + TCN) & BCI~IV-2a/2b & Strong 2025 MI baseline \\
Wang et al.~\cite{wang2025amanet} & AMANet (data augmentation + multi-scale conv) & MI & Limited-sample MI focus \\
Agarwal et al.~\cite{agarwal2025tetra} & Compact CNN for MI (tetraplegic-assistance application) & BCI~IV-2a & Small-footprint, application-oriented \\
\midrule
\multicolumn{4}{@{}l}{\textit{Transformer and hybrid CNN--Transformer models}} \\
\midrule
Zhao et al.~\cite{zhao2024ctnet} & CTNet (interleaved conv + attention) & BCI~IV-2a/2b & CNN priors stabilise small-cohort training \\
Ding et al.~\cite{ding-2024} & EEG-Deformer (dense conv-Transformer) & MI & Hierarchical local + long-range attention \\
Zhao et al.~\cite{zhao2025mscformer} & MSCFormer (parallel CNN branches + global Transformer) & BCI~IV-2a/2b & Handles inter-subject variability \\
Altaheri et al.~\cite{altaheri2025tcformer} & TCFormer (multi-kernel CNN + GQA Transformer + dilated TCN) & BCI~IV-2a/2b, HGD & Grouped-query attention back-end \\
Keutayeva et al.~\cite{keutayeva2024eegcct} & EEGCCT ($<$66K-parameter conv-Transformer) & BCI~IV-2a (LOSO) & Compact, cross-subject evaluation \\
Liao et al.~\cite{liao2025eegencoder} & EEGEncoder (dual-stream temporal--spatial block) & BCI~IV-2a & Strong 2025 MI baseline \\
Zhang et al.~\cite{zhang2023local} & Local + global convolutional attention & MI & Unified local/global attention \\
Miao et al.~\cite{satransnet2025} & SATrans-Net (Top-$K$ sparse attention) & BCI~IV-2a & Sparse attention + Grad-CAM interpretability \\
Gu et al.~\cite{cltnet2025} & CLTNet (CNN + LSTM + Transformer) & BCI~IV-2a & Sequential hybrid stack \\
Yang et al.~\cite{yang2024stmambanet} & STMambaNet (spatial/temporal Mamba) & MI & State-space alternative to attention \\
Zhang et al.~\cite{zhang2024eegmamba} & EEGMamba (bidirectional Mamba + Mixture of Experts) & MI, seizure, sleep, emotion & Multi-task linear-complexity sequence model \\
\midrule
\multicolumn{4}{@{}l}{\textit{Dual-attention and variance-sensitive feature extraction}} \\
\midrule
Du et al.~\cite{du2022dual} & Dual attentive fusion (CNN--LSTM) & BCI & Separate spatial/temporal attention; CBAM-style \\
Pan et al.~\cite{pan-2024} & DuA Transformer (decoupled temporal/channel attention) & EEG emotion & Long-term affect tracking \\
Liu et al.~\cite{liu2025dmsacnn} & DMSACNN (deep multi-scale conv + channel/spatial gating) & MI & CBAM-style gating between conv stages \\
Li et al.~\cite{li2024dual} & Dual-TSST (3 parallel Transformer branches) & MI & Temporal/spectral/spatial late fusion \\
Han et al.~\cite{han2024edpnet} & EDPNet / SST-DPN (multi-scale variance pooling) & BCI~IV-2a & Closest variance-pooling antecedent; terminal feature \\
Chen et al.~\cite{chen2024damgcn} & DAMGCN (GCN + dual attention over topology/bands) & EEG emotion & Graph-based dual attention \\
Muna et al.~\cite{sstaf2025} & SSTAF (spatial/spectral/temporal Transformer fusion) & upper-limb MI & Three-tower Transformer fusion \\
\midrule
\multicolumn{4}{@{}l}{\textit{Self-supervised pretraining and foundation models for EEG}} \\
\midrule
Kostas et al.~\cite{kostas2021bendr} & BENDR (contrastive self-supervised pretraining on raw EEG) & TUEG pretraining; downstream MI / sleep / ERP & Self-supervised pretext objective; transferable representations \\
Yang et al.~\cite{yang2023biot} & BIOT (channel- and time-tokenised biosignal Transformer) & TUEG plus multi-corpus downstream & Cross-dataset biosignal learning across heterogeneous channel sets \\
Jiang et al.~\cite{jiang2024labram} & LaBraM (Large Brain Model; vector-quantised neural codes) & 2{,}500+ hours EEG pretraining; multi-task BCI & Foundation-model approach with masked neural-code prediction \\
Wang et al.~\cite{wang2024eegpt} & EEGPT ($25$M-parameter pretrained Transformer) & mixed-corpus pretraining; downstream MI / emotion / sleep & Large-scale pretrained EEG encoder; used here under linear probing as one of the ten method families (\hyperref[subsec:ceiling]{A Signal-Limited Accuracy Ceiling}) \\
\midrule
\multicolumn{4}{@{}l}{\textit{This work}} \\
\midrule
\textbf{VC-DATNet (ours)} & Dual-attention Transformer (sensor + source space) & ToL, BISpeech, SyncSpeech (inner speech, EEG-only) & Compact ($\approx$131K-parameter) dual-attention decoder; first decoder evaluated on all three inner-speech corpora \\
\bottomrule
\end{tabularx}
\end{table}

\begin{table}[!htbp]
\centering
\scriptsize
\setlength{\tabcolsep}{3pt}
\renewcommand{\arraystretch}{1.15}
\caption{Summary of EEG-based inner-speech / imagined-speech decoding studies. ``S-D'' = subject-dependent; ``LOSO'' = leave-one-subject-out (subject-independent); ``--'' = not reported. Honest, reproducible accuracies on these tasks cluster a few points above chance ($\approx$30--36\,\% for 4-class, $\approx$15--30\,\% for 6--8-class). $^{\dagger}$\,Result is anomalously high relative to the field and, on the available evidence, attributable to within-subject over-fitting or train/test leakage (e.g.\ CSP/PLV or normalisation statistics fitted across the cross-validation split); not used as a comparison target. $^{\ddagger}$\,Competition test-set or single-subject ($n{=}1$) result, not comparable to cross-validated multi-subject accuracies.}
\label{tab:related_innerspeech}
\begin{tabularx}{\textwidth}{@{}L{2.7cm} L{1.9cm} Y L{1.9cm} L{2.4cm} Y@{}}
\toprule
\textbf{Study} & \textbf{Dataset} & \textbf{Method} & \textbf{Task (classes)} & \textbf{Acc.\ (\%)} & \textbf{Note} \\
\midrule
\multicolumn{6}{@{}l}{\textit{Thinking Out Loud (ds003626): the primary benchmark of this paper}} \\
\midrule
Nieto et al.\ 2022~\cite{nieto2022thinking}        & ToL & Dataset release; RF / CNN baselines & 4-cls directional & $\approx$25 (chance) & De facto open ToL benchmark \\
van den Berg et al.\ 2021~\cite{vandenberg2021inner} & ToL & EEGNet (2D CNN) & 4-cls directional & 29.7 & First deep-learning benchmark on ToL \\
Gasparini et al.\ 2022~\cite{Gasparini2022InnerSpeech} & ToL & SVM / XGBoost / LSTM / BiLSTM & 4-cls directional & 26.2--36.1 & BiLSTM (36.1) is the strongest prior ToL result \\
\midrule
\multicolumn{6}{@{}l}{\textit{Bimodal Inner Speech corpus / BISpeech (ds004196)}} \\
\midrule
Wilson et al.\ 2022~\cite{wilson2022bimodal}        & BISpeech & SVM / RF (EEG, fMRI) & 8-word; binary & 22.2 (8-w, EEG); 62.8 (bin., EEG) & First classification study on BISpeech \\
Simistira-Liwicki et al.\ 2023~\cite{liwicki2023openneuro,simistira2023bimodal} & BISpeech & RF / SVM; bimodal fusion & 8-word & $\approx$30 (bimodal) & BISpeech dataset descriptor \\
Wilson et al.\ 2023~\cite{wilson2023performance}    & BISpeech & EEG/fMRI early \& late fusion; EEGNet & 8-word, S-D & 14.8 (RF, EEG); 26.4 (bimodal) & Per-subject 8-word breakdown \\
Wellington et al.\ 2024~\cite{wellington2024improving} & BISpeech & Hybrid bimodal probability fusion & 8-word, S-D & $+6.0$\,pp over unimodal & Bimodal hybridisation gain \\
Qin et al.\ 2024~\cite{qin2024crossperception}      & BISpeech & ISRCP cross-perception multimodal & 4-cls soc./num. & bimodal & Multigranularity encoders + attention fusion \\
Abdalla et al.\ 2025~\cite{abdalla2025multibranch,abdalla2025plvcsp} & BISpeech & Feature-fused multi-branch deep net & 4-cls, S-D / LOSO & 50--53 (LOSO); 96--99 (S-D)$^{\dagger}$ & LOSO value is the fair target; S-D anomalous \\
Aguilar \& Grullon 2025~\cite{aguilar2025spatial}  & BISpeech & RF on band power / connectivity & binary ($n{=}1$) & 85--100$^{\ddagger}$ & Single subject only; not generalisable \\
Milyani \& Attar 2025~\cite{milyani2025innerspeech} & BISpeech & EEGNet vs.\ spectro-temporal Transformer & 8-word, LOSO & 36--38 (EEGNet); 82.4$^{\dagger}$ (Transformer) & Transformer figure outside the leak-free envelope \\
\midrule
\multicolumn{6}{@{}l}{\textit{SyncSpeech (ds006033) and other inner-speech / imagined-speech corpora}} \\
\midrule
Simistira Liwicki et al.\ 2025~\cite{simistira2025ds006033} & SyncSpeech & Dataset descriptor (EEG--fMRI) & 8-word & -- (no classifier) & Newest simultaneous EEG--fMRI corpus \\
DaSalla et al.\ 2009~\cite{dasalla2009single}      & DaSalla & CSP + nonlinear SVM & 2--3-cls vowels & 68--78 & Foundational imagined-vowel study \\
Zhao \& Rudzicz 2015~\cite{zhao2015classifying}    & KaraOne & Spectral features + DBN / SVM & binary phonological & 80--91 (binary) & KaraOne corpus; binary contrasts only \\
Coretto et al.\ 2017~\cite{coretto2017open}        & Coretto & DWT features + Random Forest & 5-vowel; 6-word & 22.3; 19.6 & Open imagined-speech database \\
Nguyen et al.\ 2018~\cite{nguyen2018inferring}     & Nguyen & Riemannian covariance + RVM & 2--3-word & 95 (2-w); 70 (3-w)$^{\ddagger}$ & Best-case binary single subjects \\
Cooney et al.\ 2020                                 & Coretto & EEGNet / Shallow / DeepConvNet & 6-word; 5-vowel & 25.0; 30.0 & Honest cross-validated DL baselines \\
BCI Competition 2020 (Track 3)                      & BCI2020 & CNN ensemble (competition winner) & 5-cls words & 82.6$^{\ddagger}$ & Competition test set; not CV-comparable \\
Lee et al.\ 2023 (Diff-E)~\cite{lee2023diffe} & BCI2020 / own & Denoising diffusion (DDPM) + classifier head & imagined-speech multi-class & competitive binary; multi-class above prior baselines & Diffusion-based representation learning for imagined-speech EEG \\
\midrule
\multicolumn{6}{@{}l}{\textit{This work}} \\
\midrule
\textbf{VC-DATNet (ours)} & ToL; BISpeech; SyncSpeech & Dual-attention Transformer (sensor + source space) & 4-cls; 8-word; binary & ToL 39.48; BISpeech 20.6 (8-w) / 59.4 (bin.); SyncSpeech 25.5 & First decoder evaluated on all three corpora; exceeds the strongest envelope-consistent baseline on ToL \\
\bottomrule
\end{tabularx}
\end{table}

\subsection*{CNN-Based EEG Decoding}

Convolutional neural networks established the foundation of deep-learning-based EEG decoding. Lawhern et al.~\cite{lawhern2018eegnet} introduced EEGNet, a compact model employing depthwise and pointwise separable convolutions to extract spatial filters and temporal dynamics with minimal parameters, and it remains the most widely referenced baseline across BCI paradigms. Schirrmeister et al.~\cite{schirrmeister2017deep} showed that ShallowConvNet and DeepConvNet, trained end-to-end on raw EEG, could match traditional filter-bank pipelines for motor imagery classification. EEG-TCNet~\cite{ingolfsson2020eeg} replaced recurrent components with temporal convolutional blocks, achieving competitive accuracy with lower latency suitable for embedded hardware.

Multi-scale receptive fields have become the dominant design choice for enriching spectral coverage within the CNN paradigm. Riyad et al.~\cite{riyad-2020} introduced Incep-EEGNet, applying inception-style parallel filters at several temporal scales. Tang et al.~\cite{tang2023motor} combined multi-scale feature extraction with frequency-domain enhancement for motor imagery. Ma et al.~\cite{ma2024attention} developed an attention-augmented CNN with multi-modal temporal fusion, explicitly capturing short- and long-range temporal structure. Xie et al.~\cite{xie2024bfatcnet} proposed BFATCNet, combining bidirectional feature pyramid attention with a temporal convolutional network to propagate multi-scale EEG features both bottom-up and top-down across frequency band representations. Wu et al.~\cite{wu2025ameegnet} introduced AMEEGNet, parallelising three EEGNet branches at different temporal scales with an efficient channel attention module, achieving 81.17\%, 89.83\%, and 95.49\% on BCI IV-2a, IV-2b, and HGD respectively. Liao et al.~\cite{liao2025ciacnet} proposed CIACNet, combining a dual-branch CNN with an improved convolutional block attention module and a temporal convolutional network, achieving 85.15\% on BCI IV-2a and 90.05\% on BCI IV-2b. Wang et al.~\cite{wang2025amanet} designed AMANet, integrating three-stage data augmentation with multi-scale temporal convolution and channel attention to address the limited-sample problem in MI-EEG.

\paragraph{Gap that motivates VC-DATNet.}
Two limitations of the CNN family directly motivate the architectural choices of this paper. First, even with multi-scale kernels and channel-attention modules, CNN-based decoders are bounded by a finite receptive field and cannot relate temporally distant neural events without substantially deeper stacks; on inner speech the relevant phenomena (alpha-band ERD envelopes spanning $\sim$500--1000\,ms, theta-locked phase episodes at $\sim$150\,ms) span both fast and slow timescales simultaneously, so a fixed-receptive-field design forces a trade-off between resolutions. VC-DATNet's combination of a multi-scale temporal front-end (up to $\approx$500\,ms, Remark~\ref{rem:receptive_field}) with a Transformer encoder having unbounded receptive field is a direct response to this trade-off. Second, every CNN architecture cited in this subsection performs feature aggregation through average pooling exclusively, discarding the band-limited variance signal that Lemma~\ref{lem:logvar_psd} identifies as the formal correlate of ERD/ERS power modulation, the dominant discriminative cue for inner speech. The mean-only pooling design point is universal in the CNN family and is what the dual statistical pooling module of VC-DATNet is designed to overturn.

\subsection*{Transformer and Hybrid CNN-Transformer Models}

Transformer architectures address the receptive field limitation through global self-attention, relating any two time points regardless of separation. Vafaei and Hosseini~\cite{vafaei2025transformers} surveyed Transformer architectures across EEG tasks, documenting consistent superiority over LSTMs for motor imagery and emotion recognition.

The prohibitive data requirements of pure Transformer training on small EEG cohorts have motivated extensive development of hybrid CNN-Transformer architectures. Zhao et al.~\cite{zhao2024ctnet} designed CTNet, interleaving local EEGNet-style convolutions with global Transformer attention, achieving 82.52\% (BCI IV-2a) and 88.49\% (BCI IV-2b) in subject-specific evaluation. Ding et al.~\cite{ding-2024} introduced EEG-Deformer, a dense convolutional Transformer using hierarchical attention for both local and long-range dependencies. Altaheri et al.~\cite{altaheri2025tcformer} proposed TCFormer, integrating a multi-kernel CNN, grouped-query attention Transformer, and dilated TCN back-end, achieving 84.79\%, 87.71\%, and 96.27\% on BCI IV-2a, IV-2b, and HGD. Zhao et al.~\cite{zhao2025mscformer} introduced MSCFormer, using multiple parallel CNN branches to address inter-subject EEG variability before global Transformer integration, reaching 82.95\% on BCI IV-2a and 88.00\% on BCI IV-2b. Keutayeva et al.~\cite{keutayeva2024eegcct} developed EEGCCT, a compact convolutional Transformer with fewer than 66K parameters achieving 70.12\% cross-subject accuracy via Leave-One-Subject-Out evaluation. Liao et al.~\cite{liao2025eegencoder} presented EEGEncoder with a Dual-Stream Temporal-Spatial Block, reaching 86.46\% subject-dependent accuracy on BCI IV-2a. Zhang et al.~\cite{zhang2023local} combined local and global convolutional attention in a unified framework for motor imagery. SATrans-Net~\cite{satransnet2025} replaced dense self-attention with Top-K sparse attention, achieving 84.72\% on BCI IV-2a with reduced computational cost and Grad-CAM cortical interpretability. CLTNet~\cite{cltnet2025} further combined CNN, LSTM, and Transformer in sequence, attaining 83.02\% on BCI IV-2a. Collectively, these works confirm that convolutional inductive biases remain essential for Transformer-based EEG decoding on small datasets.

State-space models have recently emerged as an alternative to attention for long-sequence modelling. Yang and Jia~\cite{yang2024stmambanet} proposed STMambaNet, leveraging the Mamba architecture with spatial and temporal Mamba encoders for linear-complexity sequence modelling of MI-EEG signals. Zhang et al.~\cite{zhang2024eegmamba} introduced EEGMamba, combining bidirectional Mamba with a Mixture of Experts for multi-task EEG classification across motor imagery, seizure detection, sleep staging, and emotion recognition.

\paragraph{Gap that motivates VC-DATNet.}
Three structural commitments are common to every Transformer and hybrid architecture cited above, and they persist in the most recent 2024--2025 work. VC-DATNet departs from each of them deliberately. (i) Self-attention is applied uniformly: every time step and every electrode receives equal a-priori capacity, which on a spatially diffuse and temporally transient signal such as inner speech dilutes the informative subspace. VC-DATNet replaces uniform self-attention with selectively gated self-attention (\hyperref[subsec:dual_attention]{Dual Attention Transformer Encoder}) that suppresses uninformative content before softmax aggregation, a property formalised in Proposition~\ref{prop:simplex}. (ii) Feature aggregation is exclusively mean-based: none of the cited Transformer-based models pairs the attention pathway with a parallel variance representation, which Lemma~\ref{lem:logvar_psd} shows is the natural primitive for ERD/ERS-driven decoding. (iii) Evaluation is overwhelmingly on motor-imagery benchmarks (BCI IV-2a/2b, HGD); the adequacy of the resulting architectural priors for inner-speech decoding is therefore not tested. VC-DATNet evaluates its dual-attention encoder on three inner-speech corpora (ToL, BISpeech, and SyncSpeech; \hyperref[subsec:innerspeech_results]{Inner-Speech Decoding Across Three Datasets}), precisely because the architectural commitments of MI-trained Transformers are unlikely to transfer when the discriminative cue moves from amplitude to second-order statistics.

\subsection*{Dual Attention and Variance-Sensitive Feature Extraction}

Selective attention mechanisms have been actively explored to direct model capacity towards physiologically informative EEG regions, and three families have crystallised in the literature. Each is structurally different from the dual-attention design proposed in VC-DATNet, and the differences are worth making explicit because they are the source of VC-DATNet's contribution claim.

\paragraph{Family 1: CBAM-style gating placed outside the attention block.}
Du et al.~\cite{du2022dual} introduced a dual attentive fusion framework that applies separate spatial and temporal attention modules to a CNN--LSTM backbone, gating the input feature map before classification. Liu et al.~\cite{liu2025dmsacnn} adopted a similar pattern in DMSACNN, inserting channel and spatial attention gating between convolutional stages, and Chen et al.~\cite{chen2024damgcn} combined graph convolutional networks with dual attention over channel topology and frequency bands for emotion recognition. These designs are CBAM~\cite{woo2018cbam}-style gating mechanisms placed outside a self-attention block (most often without one); they treat temporal and channel axes symmetrically, and they do not interact with a Transformer's $\mathbf{Q}/\mathbf{K}/\mathbf{V}$ projections.

\paragraph{Family 2: Decoupled-axis attention applied post-softmax or to projections.}
Pan et al.~\cite{pan-2024} developed the DuA Transformer for long-term EEG emotion analysis, showing that decoupling temporal and channel attention substantially improved continuous tracking of affective state. Pan et al.\ apply the two attention pathways to the input embedding (and to the output of attention via residual fusion), not to the value tensor of multi-head attention itself. The temporal and channel axes are treated symmetrically, and the resulting model has no second statistical stream.

\paragraph{Family 3: Parallel single-attention branches over separate representations.}
Li et al.~\cite{li2024dual} introduced Dual-TSST, processing temporal, spectral, and spatial representations through three parallel Transformer branches that are concatenated at a late fusion stage. Xie et al.~\cite{sstaf2025} adopted the same pattern in SSTAF, fusing spatial, spectral, and temporal Transformer encoders for upper-limb motor imagery. These models pluralise the Transformer rather than the gating, and within each branch standard self-attention treats all positions uniformly.

\paragraph{Variance pooling as a separate strand.}
Independently of the dual-attention literature, second-order feature aggregation has been recognised as physiologically motivated for ERD/ERS-driven decoding. Han et al.~\cite{han2024edpnet} introduced EDPNet, whose multi-scale variance pooling (MVP) module captures spectral power variation at multiple temporal scales without trainable parameters in the pooling stage, reaching 84.11\,\% on BCI IV-2a. MVP is a terminal feature extractor whose output is fed directly to the classifier; it does not condition any attention or Transformer mechanism, and it is not coupled to a parallel mean-activity stream within an attention pathway.

\paragraph{Positioning of VC-DATNet.}
VC-DATNet differs structurally from each of these families along three axes simultaneously, and it is the conjunction of the three, not any single one, that defines the architectural contribution.

(i) Pre-softmax value-tensor gating. The temporal and spatial gates are applied to $\mathbf{V}$ before the standard $\mathrm{softmax}(\mathbf{Q}\mathbf{K}^{\top}/\sqrt{d_k})$ aggregation, so the query--key similarity structure is preserved while the content routed through that structure is selectively suppressed. Family 1 places gating outside the attention block; Family 2 places it on residual streams or post-attention; Family 3 has no V-conditioning at all.

(ii) Asymmetric gate derivation. The temporal gate is produced by an MLP applied to $\mathbf{V}$ at full sequence resolution (per-timestep), while the spatial gate is produced by an MLP applied to a sequence-mean-pooled global summary $\bar{\mathbf{V}}$. This asymmetry mirrors EEG physiology: temporal selectivity must respect the millisecond-scale dynamics of event-related desynchronisation, whereas electrode informativeness is comparatively stationary across a trial and benefits from sequence-level integration. Existing dual-attention works (Families 1--3) treat the two axes symmetrically.

(iii) Variance-coupled dual-stream attention. The dual-attention encoder is instantiated twice in parallel, once over a mean-pooled stream and once over a log-variance-pooled stream, and the streams are fused only at the classifier head. EDPNet's MVP is the closest variance-pooling antecedent but consumes variance as a final feature rather than as a representation that conditions attention. To our knowledge, no prior architecture gates Transformer attention with a variance-encoded stream, a property that, by Lemma~\ref{lem:logvar_psd}, is well-matched to inner speech, where the discriminative signal lives almost entirely in second-order statistics (\hyperref[sec:introduction]{Introduction}).

We do not claim that any individual gating, attention, or pooling component is new in isolation. The architecture's role in this paper is as a consistent, reproducible instrument applied uniformly across the three inner-speech datasets, and the specific conjunction described above (pre-softmax V-gating, asymmetric per-axis gate derivation, and a dual-stream parallel encoding) is documented here to support reproduction rather than as a primacy claim.

\subsection*{Inner Speech EEG Decoding}

Motor imagery is the dominant evaluation paradigm in BCI literature, yet inner speech is the more cognitively expressive and clinically relevant target for severely motor-impaired populations~\cite{nieto2022thinking}. Inner speech decoding, the classification of silent verbal commands from EEG, presents fundamentally different signal characteristics: neural correlates are more spatially diffuse, signal-to-noise ratios are substantially lower, and within-subject variability is higher than in overt movement preparation~\cite{nieto2022thinking, innerspeech_future2025, imaginedspeech_review2025}.

Three open datasets are used in this paper. Thinking Out Loud (ToL)~\cite{nieto2022thinking} provides 4-class directional inner speech (up/down/left/right) from 10 subjects. Gasparini et al.~\cite{Gasparini2022InnerSpeech} established the early CNN and RNN baselines on this corpus, and a recent systematic review~\cite{imaginedspeech_review2025} identifies cross-subject generalisation and low signal amplitude as the principal open challenges for the paradigm. The Bimodal Inner Speech corpus (BISpeech; OpenNeuro ds004196)~\cite{liwicki2023openneuro,simistira2023bimodal} provides 8-class semantic word-level inner speech (4 social, 4 numerical) from 4 subjects with simultaneously recorded EEG and fMRI; we use only its EEG modality. The Synchronous EEG--fMRI Inner Speech corpus (SyncSpeech; OpenNeuro ds006033)~\cite{simistira2025ds006033} is a third, newer EEG--fMRI inner-speech corpus with the same 8-word vocabulary as BISpeech but a different cohort of 3 subjects, and (in contrast to BISpeech) records the two modalities simultaneously. We decode it here for the first time.

The classification literature on BISpeech is small but growing. Wilson et al.~\cite{wilson2022bimodal,wilson2023performance} reported subject-dependent SVM, random forest, and EEGNet baselines (8-word EEG-only mean 22.19\,\%; bimodal late fusion 26.41\,\%). Wellington et al.~\cite{wellington2024improving} reported a hybrid bimodal model with $+6.0$\,pp uplift over the unimodal baseline. Qin et al.~\cite{qin2024crossperception} introduced the bimodal ISRCP cross-perception model (47.20\,\% on 4-class social, 36.50\,\% on 4-class numerical). Abdalla et al.~\cite{abdalla2025multibranch,abdalla2025plvcsp} reported subject-independent EEG-only deep-learning results (4-class: 50--53\,\%) alongside anomalously high subject-dependent results (96--100\,\%). Milyani and Attar~\cite{milyani2025innerspeech} compared EEGNet against a wavelet-feature spectro-temporal Transformer on the EEG modality of BISpeech in an 8-word leave-one-subject-out evaluation, reporting $82.4\,\%$ for the Transformer against about $36\,\%$ for EEGNet. The Abdalla subject-dependent values and the Milyani Transformer value both exceed every other EEG-only result on this corpus by more than $50$ percentage points and lie well outside any plausible leak-free signal-to-noise envelope. We set them aside and take the envelope-consistent figures as the credible baselines: Wilson's $22.19\,\%$ on 8-word and Abdalla's $50$--$53\,\%$ on the 4-class subtasks. We return to this point in \hyperref[subsec:innerspeech_results]{Inner-Speech Decoding Across Three Datasets}.

\paragraph{Gap that motivates VC-DATNet.}
Three observations follow from the literature reviewed above. First, the architectures surveyed in the preceding subsections (CTNet, EEG-Deformer, MSCFormer, TCFormer, EEGEncoder, CIACNet, AMEEGNet, EDPNet, SSTAF) are validated almost exclusively on motor imagery; not one is evaluated on an inner-speech corpus, so the adequacy of their architectural priors for inner speech remains untested. Second, the inner-speech studies cited above (Table~\ref{tab:related_innerspeech}) are each reported on a single corpus under a single evaluation protocol, so genuine, transferable findings cannot be separated from dataset- or model-specific artefacts. Third, no published architecture is evaluated on more than one inner-speech corpus, so within-paradigm cross-corpus replication, the standard form of external validity, has not been attempted. This paper addresses all three: one fixed architecture and protocol (Table~\ref{tab:hyperparams}) is applied to three independent inner-speech datasets, ToL (\hyperref[subsec:innerspeech_results]{Inner-Speech Decoding Across Three Datasets}), the EEG modality of BISpeech (\hyperref[subsec:innerspeech_results]{Inner-Speech Decoding Across Three Datasets}), and SyncSpeech, with each dataset additionally decoded from two signal representations (sensor- and source-space) so that the comparison is controlled rather than confounded by model selection.

\section*{Methodology}\phantomsection \label{subsec1}
This section specifies the datasets, the single preprocessing pipeline (artefact removal, band-pass filtering, channel selection, epoching), and the VC-DATNet architecture (multi-scale temporal convolution, depthwise spatial convolution, dual statistical pooling, and a two-stream dual-attention Transformer encoder), in that order. All three datasets pass through the same pipeline and are decoded by the same architecture under one training recipe (Table~\ref{tab:hyperparams}), so that any difference reported in Results is attributable to the data or the signal representation rather than to model selection.

\subsection*{Datasets}
VC-DATNet is evaluated on three independent public inner-speech EEG corpora: Thinking Out Loud (ToL; primary), the EEG modality of BISpeech, and SyncSpeech (the latter decoded here for the first time). The three differ in cohort size, channel count, stimulus type, and acquisition hardware (Table~\ref{tab:datasets}); behaviour that recurs across all three is therefore evidence of a paradigm-level effect rather than a single-dataset artefact (\hyperref[subsec:innerspeech_results]{Inner-Speech Decoding Across Three Datasets}).

% \begin{table}[!htbp]
% \centering
% \caption{The three public inner-speech EEG datasets used in this study, all obtained from OpenNeuro. The EEG modality only is used; for BISpeech and SyncSpeech the simultaneously recorded fMRI is not used. All datasets are evaluated under within-subject five-fold cross-validation. ``Channels'' is the recorded scalp-electrode count; the sensor- and source-space configurations actually decoded are described in \hyperref[subsec2]{Data Preprocessing}.}
% \label{tab:datasets}
% %
% \begin{tabular}{l|c|c|c|c|c|l}
% \toprule
% \textbf{Dataset} & \textbf{Year} & \textbf{Subjects} & \textbf{Channels} & \textbf{Classes} & \textbf{Trials} & \textbf{Stimuli} \\
% \midrule
% Thinking Out Loud (ToL; ds003626)~\cite{nieto2022thinking} & 2022 & 10 & 128 & 4 & $\sim$50/class ($\sim$200/subject) & directional words (up/down/left/right) \\
% BISpeech (ds004196)~\cite{liwicki2023openneuro,simistira2023bimodal} & 2023 & 4 & 64 & 8 & 40/word & semantic words (4 social, 4 numerical) \\
% SyncSpeech (ds006033)~\cite{simistira2025ds006033} & 2025 & 3 & 64 & 8 & 40/word & semantic words (4 social, 4 numerical) \\
% \bottomrule
% \end{tabular}%

% \end{table}

\begin{table}[!htbp]
\centering
\caption{The three public inner-speech EEG datasets used in this study, all
obtained from OpenNeuro. Only the EEG modality is used; the simultaneously
recorded fMRI data in BISpeech and SyncSpeech are not analysed. All datasets
are evaluated using within-subject five-fold cross-validation. Channels
denotes the number of recorded scalp electrodes; the decoded sensor- and
source-space configurations are described in
\hyperref[subsec2]{Data Preprocessing}.}
\label{tab:datasets}

\small
\setlength{\tabcolsep}{3pt}
\renewcommand{\arraystretch}{1.15}

\begin{tabularx}{\textwidth}{
    @{}
    >{\raggedright\arraybackslash}p{2.7cm}
    c c c c
    >{\centering\arraybackslash}p{1.8cm}
    Y
    @{}
}
\toprule
\textbf{Dataset} &
\textbf{Year} &
\textbf{Subj.} &
\textbf{Ch.} &
\textbf{Classes} &
\textbf{Trials} &
\textbf{Stimuli} \\
\midrule

Thinking Out Loud
(ToL; ds003626)~\cite{nieto2022thinking}
&
2022 & 10 & 128 & 4
&
$\sim$50/class
($\sim$200/subject)
&
Directional words:
up, down, left, and right
\\

BISpeech
(ds004196)~\cite{liwicki2023openneuro,simistira2023bimodal}
&
2023 & 4 & 64 & 8
&
40/word
&
Semantic words:
four social and four numerical
\\

SyncSpeech
(ds006033)~\cite{simistira2025ds006033}
&
2025 & 3 & 64 & 8
&
40/word
&
Semantic words:
four social and four numerical
\\

\bottomrule
\end{tabularx}
\end{table}

\textbf{Thinking Out Loud (ToL): 4-Class Inner Speech (primary dataset).}

The Thinking Out Loud corpus (OpenNeuro ds003626)~\cite{nieto2022thinking} is the principal benchmark used in this work. It comprises EEG recordings from 10 subjects who imagined speaking four directional commands (``Arriba''/up, ``Abajo''/down, ``Derecha''/right, ``Izquierda''/left). Of the four task conditions in the original Nieto et al.\ corpus (pronounced, inner, visualised, and stimulus), this paper uses only the inner-speech condition, with each subject contributing on the order of $200$ usable inner-speech trials (approximately $50$ trials per directional class, balanced; the exact count per subject ranges from $180$ to $240$ depending on artefact rejection). Data were acquired using a 136-channel BioSemi ActiveTwo system, of which 128 were active scalp EEG electrodes and 8 external channels for EOG and EMG monitoring. ToL is the only public inner speech EEG benchmark with sufficient trial count and subject coverage to support end-to-end deep-learning evaluation, and is therefore the de facto standard for the paradigm~\cite{Gasparini2022InnerSpeech,imaginedspeech_review2025}. Compared with the motor-imagery paradigm that the architectural literature is primarily benchmarked on, the discriminative neural signal in ToL is spatially diffuse, lower in amplitude, and more variable across trials, the three properties that motivate VC-DATNet's architectural choices (\hyperref[sec:introduction]{Introduction}).

\textbf{Liwicki Bimodal Inner Speech (BISpeech): EEG-only configuration (secondary inner speech dataset).}

The Bimodal Inner Speech corpus (hereafter \emph{BISpeech}; OpenNeuro ds004196)~\cite{liwicki2023openneuro,simistira2023bimodal} provides bimodal EEG and fMRI recordings from four subjects (sub-01, sub-02, sub-03, sub-05; sub-04 was excluded by the dataset authors due to EEG signal degradation) performing an inner-speech task on eight stimulus words organised into two semantic categories: four social terms (\textit{child}, \textit{daughter}, \textit{father}, \textit{wife}) and four numerical terms (\textit{three}, \textit{four}, \textit{six}, \textit{ten}), with 40 trials per word per modality (320 trials total per subject). EEG was recorded with a 64-channel BioSemi ActiveTwo system at 512\,Hz; after rejecting noisy electrodes, $48$ channels common to all subjects were retained for decoding. We include this dataset as a secondary inner speech benchmark to test whether VC-DATNet's decoding behaviour replicates on a structurally different inner speech corpus: one with smaller cohort size (4 vs.\ 10 subjects), denser channel coverage (48 vs.\ 22), and \emph{semantic} stimuli rather than the directional commands of ToL. Following the EEG-only positioning of this paper, we use only the EEG modality of BISpeech; bimodal EEG--fMRI methods reported in~\cite{wilson2022bimodal,wilson2023performance,wellington2024improving,qin2024crossperception} are noted for context but are out of scope for direct comparison.

\textbf{Synchronous EEG--fMRI Inner Speech (SyncSpeech; OpenNeuro ds006033): 8-class semantic inner speech, third inner-speech corpus.}

SyncSpeech~\cite{simistira2025ds006033} is an inner-speech EEG corpus recorded with simultaneous EEG and fMRI from three subjects, using a 64-channel MR-compatible BrainCap MR cap. Three subjects silently articulated an eight-word vocabulary of four social and four numerical terms, the same semantic structure as BISpeech. EEG was acquired with an MR-compatible cap; we use only the EEG modality. To our knowledge no decoding result has previously been reported on this corpus. We include it as an independent third test of whether the behaviour observed on ToL and BISpeech, in particular the sensor/source two-regime effect, generalises to a further inner-speech dataset recorded with different hardware and a smaller cohort.

To make the comparison with prior work as transparent as possible, all three datasets are processed through an identical preprocessing pipeline (\hyperref[subsec2]{Data Preprocessing}) and decoded by one architecture and one training recipe; every training and capacity hyperparameter (learning rate, batch size, encoder depth, embedding dimension, attention heads, kernel widths, dropout, epochs) is held fixed across all three datasets, and the only quantities that vary are the two input dimensions the data itself fixes, the EEG channel count and the output-class count; the full configuration is reported in Table~\ref{tab:hyperparams}. The three inner-speech corpora together provide a within-paradigm cross-corpus replication: behaviour that recurs across ToL, BISpeech, and SyncSpeech, which differ in cohort size, stimulus type, channel count, and acquisition hardware, is unlikely to be a single-dataset artefact.

\subsection*{Data Preprocessing}\phantomsection \label{subsec2}

A single preprocessing pipeline, implemented in MNE-Python~\cite{Gramfort2013} and identical across the three datasets, converts the raw recordings into the fixed-shape epochs the network consumes. All three datasets are resampled to $256$\,Hz; each task-relevant imagery interval is epoched to a $2.5$\,s window ($\approx 640$ samples), which is the input length specified in Table~\ref{tab:hyperparams}. Thinking Out Loud uses a $136$-channel BioSemi montage; the eight non-scalp (EXG) channels and any noisy electrodes are dropped, and sensor-space decoding uses the $22$ inner-speech-relevant electrodes covering frontal, temporal, and parietal regions implicated in inner-speech production. BISpeech is recorded with $64$ EEG channels (we retain the $48$ usable after bad-channel rejection); SyncSpeech is recorded with a $64$-channel MR-compatible cap. The same epochs are additionally projected to a cortical source-space configuration via an sLORETA inverse solution and averaged within $20$ language-network ROIs of the Desikan--Killiany atlas (\hyperref[subsec:sourcespace]{Source-Space Projection}); this is the second signal representation used in Results.

\subsubsection*{Raw Data Inspection and Event Annotation}\phantomsection \label{subsec3}

Each raw recording is loaded and visually inspected for gross artefact or channel failure before processing. Trial onsets and durations are read from the dataset-supplied TSV event files, converted from milliseconds to sample indices using each dataset's native sampling rate, and assigned a unique event label per stimulus (e.g., ``child,'' ``father,'' ``daughter''). These labels are attached as annotations to the continuous EEG and drive the epoching stage below.

\subsubsection*{Channel Configuration and Referencing}\phantomsection \label{subsec4}

Each dataset is configured to its native standard montage and reduced to scalp EEG by removing non-scalp channels (e.g., EXG1--EXG6). An average reference is then applied to the remaining channels, normalising voltage across the scalp so that no electrode is implicitly privileged in downstream analysis. Denoting the signal on channel \(i\) as \(x_i(t)\), the re-referenced signal \(x_i'(t)\) is
\[
  x_i'(t)\;=\;x_i(t)\;-\;\frac{1}{N}\sum_{j=1}^{N}x_j(t),
\]
where \(N\) is the total number of channels \cite{dong-2021}. The positions of the 22 retained channels for the Thinking Out Loud sensor-space configuration are shown in Supplementary Figure~S1.

\subsubsection*{Noise Reduction through Filtering}\phantomsection \label{subsec5}

Three filter stages target the dominant noise sources in scalp EEG. A notch filter at $50$\,Hz and its harmonics ($100$ and $150$\,Hz) removes mains line interference; a $0.5$\,Hz high-pass eliminates slow drift; and a $4$--$40$\,Hz band-pass retains the frequency range relevant to inner-speech decoding while suppressing the remaining out-of-band noise~\cite{Widmann2015}. If \(X(f)\) is the Fourier transform of the raw signal \(x(t)\) and \(H(f)\) the filter's frequency response, the filtered signal \(y(t)\) is
\[
  y(t)
  = \mathcal{F}^{-1}\bigl\{H(f)\,X(f)\bigr\},
\]
where
\[
  H(f) =
    \begin{cases}
      1, & 0.5 \le |f| \le 40,\\
      0, & \text{otherwise.}
    \end{cases}
\]

\subsubsection*{Artifact Removal Using ICA}\phantomsection \label{subsec6}

Independent Component Analysis (ICA) suppresses residual physiological artefacts, principally ocular activity. In the absence of dedicated EOG channels in the public recordings, the Fp1 electrode serves as the ocular-artefact surrogate; ICA components with topographies and time courses consistent with eye blinks or saccades are identified and removed using the MNE-Python implementation~\cite{ILLE2024149, Delorme2004}.

\subsubsection*{Epoching the Data}\phantomsection \label{subsec7}

The cleaned signal is segmented into fixed-length trials. Each epoch is resampled to $256$\,Hz and cropped to a $2.5$\,s window ($\approx 640$ samples) aligned to imagery onset, with baseline correction over the $-0.2$ to $0$\,s pre-onset interval~\cite{Kappenman2021}; this is the input length the network expects (Table~\ref{tab:hyperparams}). On Thinking Out Loud we compare two channel configurations:

\begin{itemize}
    \item The full $128$-channel scalp montage (used by the cross-architecture bench in Table~\ref{tab:ceiling_full}).
    \item A $22$-channel a-priori subset covering the inner-speech-relevant frontal, temporal, and parietal regions; this is the VC-DATNet headline configuration on ToL.
\end{itemize}

BISpeech uses the $48$ channels remaining after bad-channel rejection; SyncSpeech uses its $64$-channel MR-compatible montage. The $22$-channel subset on ToL was selected a priori from the regions implicated in inner-speech production by prior ERP work~\cite{xu-2024,Kappenman2021,PerroneBertolotti2014}; this choice is documented for transparency but does not enter the headline accuracy numbers.

The full preprocessing pipeline (filter stages, ICA artefact removal, montage selection, and epoching) delivers fixed-shape epochs that retain the inner-speech-relevant signal while suppressing the dominant scalp-EEG noise sources, and is applied unchanged across all three datasets and both signal representations used in the Results.

\subsection*{Source-Space Projection}\phantomsection \label{subsec:sourcespace}

For each dataset we additionally decode from a cortical source-space projection of the same epoched EEG. The projection is computed with the MNE-Python implementation~\cite{Gramfort2013} of the standardised low-resolution brain electromagnetic tomography (sLORETA) inverse method~\cite{pascualmarqui2002sloreta}; the pipeline below is the same for all three datasets.

\paragraph{Forward model.}
We use the FreeSurfer \texttt{fsaverage} template subject as the head model, which provides a standard cortical surface, source space, and boundary-element model (BEM) of the head. The cortical source space is the ico-5 mesh ($\approx$10\,242 sources per hemisphere). A three-layer BEM (scalp, outer skull, inner skull) computes the volume conductor and the lead-field matrix from every source location to every scalp electrode. Each dataset's electrode positions are placed on the template head using the standard montage corresponding to its recording cap (BioSemi 128 for ToL; manufacturer-supplied positions for BISpeech and SyncSpeech). EEG is referenced to the average across electrodes prior to source localisation.

\paragraph{Noise covariance and inverse operator.}
The inner-speech protocols do not provide a clean pre-stimulus baseline of sufficient length for a stable empirical noise covariance estimate, so we use an ad-hoc identity-scaled covariance appropriate for the sensor type (\texttt{mne.make\_ad\_hoc\_cov}). The inverse operator is constructed from the forward solution and the noise covariance with loose source-orientation constraint ($\mathrm{loose} = 0.2$) and depth weighting ($\mathrm{depth} = 0.8$), the MNE-Python defaults for cortically constrained source localisation. The sLORETA regularisation parameter is $\lambda^{2} = 1/\mathrm{SNR}^{2}$ with assumed $\mathrm{SNR} = 3$, giving $\lambda^{2} \approx 0.111$.

\paragraph{Region-of-interest selection.}
Source time courses are summarised over $20$ regions of interest (ROIs) of the Desikan--Killiany cortical atlas~\cite{desikan2006automated}, chosen \emph{a priori} to cover the cortical language network and adjacent motor/premotor regions implicated in inner speech~\cite{PerroneBertolotti2014,nieto2022thinking}: bilateral inferior frontal gyrus (Broca's area: \emph{pars opercularis}, \emph{pars triangularis}, and \emph{pars orbitalis} in the left hemisphere; \emph{pars opercularis} in the right); bilateral superior temporal gyrus and banks of the superior temporal sulcus, with the transverse temporal gyrus (auditory cortex) bilaterally and the middle temporal gyrus bilaterally for semantic processing; bilateral precentral and postcentral gyri, together with the paracentral lobule, for the motor-speech network; and the bilateral supramarginal gyrus together with the left inferior parietal lobule for the parietal language regions. The exact ROI list, the source-space code, and the per-dataset preprocessing scripts are released alongside the model.

\paragraph{ROI time courses.}
For each epoch, the sLORETA inverse operator is applied to obtain a cortical source estimate, and the time course of each ROI is extracted by averaging the source current density across the vertices within that ROI's label (\texttt{mne.extract\_label\_time\_course} with \texttt{mode = "mean"}). The result is a tensor of shape $(B, 20, T)$ that replaces the sensor-space tensor $(B, C, T)$ for downstream training; the architecture and training recipe (Table~\ref{tab:hyperparams}) are otherwise unchanged. The two representations (sensor- and source-space) are decoded independently with identical hyperparameters, and the resulting accuracies are compared per subject to identify the two-regime effect reported in \hyperref[subsec:innerspeech_results]{Inner-Speech Decoding Across Three Datasets}.

\subsection*{VC-DATNet Architecture}

\begin{figure}[H]
    \centering
    \includegraphics[width=1\linewidth]{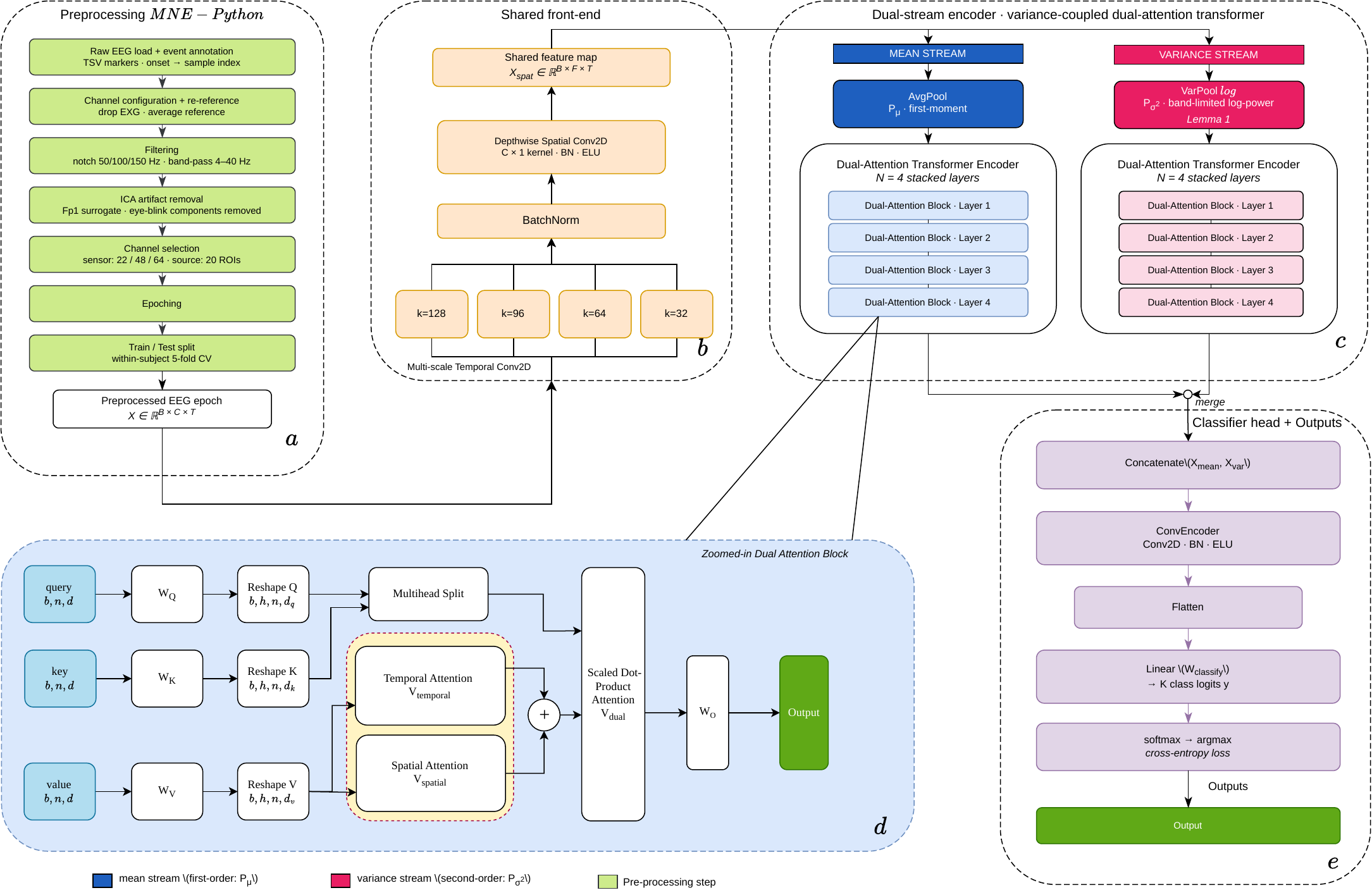}
    \caption{End-to-end pipeline and architecture of VC-DATNet. Panel (a) covers preprocessing (\hyperref[subsec2]{Data Preprocessing}): raw-EEG loading and event annotation, channel configuration and average re-referencing, $4$--$40$\,Hz band-pass filtering, ICA artefact removal, channel selection, and epoching to a $2.5$\,s window at $256$\,Hz. The output is a preprocessed epoch $\mathbf{X}\in\mathbb{R}^{B\times C\times T}$, where $C$ is the dataset-determined sensor-channel count or, for the source-space configuration (\hyperref[subsec:sourcespace]{Source-Space Projection}), $20$ language-network regions of interest. Panel (b) shows the shared front-end. A multi-scale temporal Conv2D with four parallel kernels of $\{32,64,96,128\}$ samples (Eqs.~\ref{eq:Xi}--\ref{eq:Xtemp}) is followed by batch normalisation and a depthwise spatial Conv2D ($C\times1$, BatchNorm, ELU; Eq.~\ref{eq:Xspat}), producing the feature map $\mathbf{X}_{\mathrm{spat}}\in\mathbb{R}^{B\times F\times T}$. Panel (c) shows the dual-stream encoder. The feature map $\mathbf{X}_{\mathrm{spat}}$ is pooled in parallel into a mean stream ($\mathcal{P}_\mu$, Eq.~\ref{eq:avgpool}) and a clamped log-variance stream ($\mathcal{P}_{\sigma^2}$, Eq.~\ref{eq:varpool}); the latter is a per-window estimator of band-limited log-power (Lemma~\ref{lem:logvar_psd}). Each stream is processed independently by a $4$-layer dual-attention Transformer encoder (Eqs.~\ref{eq:V_temporal}--\ref{eq:dual_attn_softmax}). Panel (d) is a zoomed view of one dual-attention block. The spatial and channel gates are derived from the value tensor at full sequence resolution and from a sequence-mean summary respectively. They are added to form $\mathbf{V}_{\mathrm{dual}}$ before the scaled dot-product attention $\mathrm{softmax}(\mathbf{Q}\mathbf{K}^{\top}/\sqrt{d_k})$, which leaves the query--key similarity structure intact (Proposition~\ref{prop:simplex}); the result is projected by $\mathbf{W}_O$. Panel (e) is the classifier head and outputs. The two transformed streams are concatenated (Eq.~\ref{eq:concat}), integrated by a small convolutional encoder (Conv2D, BatchNorm, ELU), flattened, and projected to $K$ class logits (Eq.~\ref{eq:logits}). Softmax and arg-max produce the class-probability vector $\mathrm{softmax}(\mathbf{y})\in\Delta^{K-1}$ and the predicted label $\arg\max_k y_k$. Training is end-to-end with cross-entropy (Eq.~\ref{eq:ce_loss}). The network corresponds to the operator composition of Definition~\ref{def:datnet}.}
    \label{fig:overview}
\end{figure}

VC-DATNet is a hybrid deep-learning architecture designed to model the spatiotemporal and second-order statistical structure of EEG signals for inner-speech decoding. To make the architectural claims of the paper precise, we first state the notation, then define VC-DATNet as a composition of typed operators.

\subsubsection*{Notation}

Throughout this section, $B$ denotes the batch size, $C$ the number of EEG channels, $T$ the number of temporal samples per epoch, $D$ the model embedding dimension, $H$ the number of attention heads (with per-head dimension $d_k = D/H$), $L$ the temporal length after pooling, and $K$ the number of classes. The principal tensors are summarised in Table~\ref{tab:notation}; for a tensor $\mathbf{Z}$ we write $\mathbf{Z}_{i,:,:}$ for the $i$-th batch slice.

\begin{table}[!htbp]
\centering
\caption{Principal symbols and shapes used throughout the architecture description.}
\label{tab:notation}
\begin{tabular}{l|l|l}
\toprule
\textbf{Symbol} & \textbf{Shape} & \textbf{Meaning} \\
\midrule
$\mathbf{X}$                  & $\mathbb{R}^{B \times C \times T}$         & Preprocessed EEG epoch \\
$\mathbf{X}_{\text{temp}}$    & $\mathbb{R}^{B \times F_t \times C \times T}$ & Multi-scale temporal features \\
$\mathbf{X}_{\text{spat}}$    & $\mathbb{R}^{B \times F \times 1 \times T}$    & Depthwise-spatial features \\
$\mathbf{P}_{\mu}, \mathbf{P}_{\sigma^2}$ & $\mathbb{R}^{B \times F \times L}$ & Mean / log-variance pooled streams \\
$\mathbf{Q}, \mathbf{K}, \mathbf{V}$ & $\mathbb{R}^{B \times N \times D}$ & Query / key / value tensors ($N{=}L$) \\
$\mathbf{g}_t \in [0,1]^{B \times N \times D}$ & per-token & Temporal gate (per-timestep) \\
$\mathbf{g}_s \in [0,1]^{B \times 1 \times D}$ & broadcast & Spatial gate (sequence-mean MLP) \\
$\mathbf{V}_{\text{dual}}$    & $\mathbb{R}^{B \times N \times D}$         & Gated value tensor (Eq.~\ref{eq:V_dual}) \\
$\mathbf{y}$                  & $\mathbb{R}^{B \times K}$                  & Class logits \\
\bottomrule
\end{tabular}
\end{table}

\begin{definition}[VC-DATNet]
\label{def:datnet}
Given the operators
\begin{align*}
&\mathcal{T}     &&\colon \mathbb{R}^{B \times C \times T} \to \mathbb{R}^{B \times F_t \times C \times T} && \text{(multi-scale temporal conv)} \\
&\mathcal{S}     &&\colon \mathbb{R}^{B \times F_t \times C \times T} \to \mathbb{R}^{B \times F \times 1 \times T} && \text{(depthwise spatial conv)} \\
&\mathcal{P}_{\mu}, \mathcal{P}_{\sigma^2} &&\colon \mathbb{R}^{B \times F \times 1 \times T} \to \mathbb{R}^{B \times F \times L} && \text{(mean / log-variance pooling)} \\
&\mathcal{A}     &&\colon \mathbb{R}^{B \times N \times D} \to \mathbb{R}^{B \times N \times D} && \text{(dual-attention encoder)} \\
&\mathcal{H}     &&\colon \mathbb{R}^{B \times N \times D} \times \mathbb{R}^{B \times N \times D} \to \mathbb{R}^{B \times K} && \text{(classifier head)}
\end{align*}
the VC-DATNet predictor $f_{\theta}\colon \mathbb{R}^{B \times C \times T} \to \mathbb{R}^{B \times K}$ is the composition
\begin{equation}
f_{\theta}(\mathbf{X}) = \mathcal{H}\bigl(\, \mathcal{A}(\mathcal{P}_{\mu} \circ \mathcal{S} \circ \mathcal{T})(\mathbf{X}),\ \mathcal{A}(\mathcal{P}_{\sigma^2} \circ \mathcal{S} \circ \mathcal{T})(\mathbf{X}) \,\bigr),
\label{eq:datnet_def}
\end{equation}
where the two arguments of $\mathcal{H}$ are the mean and log-variance streams produced by independent invocations of the dual-attention encoder $\mathcal{A}$, and $\theta$ collects all learnable parameters. The remainder of this section instantiates each operator and proves a structural property of $\mathcal{A}$ (Proposition~\ref{prop:simplex}).
\end{definition}

Figure~\ref{fig:overview} lays out the constituent components of VC-DATNet, which we now describe in turn. Each subsection closes with an equation, a remark on receptive field or complexity, or a formal property.

\paragraph{Outline of the operators.}
$\mathcal{T}$ is a multi-scale temporal convolution with four parallel 2D kernels of length $k_i \in \{32, 64, 96, 128\}$ samples (\hyperref[subsubsec:multiscale_temporal]{Multi-scale Temporal Convolution}, Eqs.~\ref{eq:Xi}--\ref{eq:Xtemp}); $\mathcal{S}$ is a depthwise-spatial convolution with batch normalisation and ELU non-linearity (\hyperref[subsubsec:spatial_conv]{Spatial Convolution}, Eq.~\ref{eq:Xspat}); $\mathcal{P}_{\mu}$ and $\mathcal{P}_{\sigma^2}$ are the mean and clamped log-variance pooling operators of \hyperref[subsec:dual_pooling]{Dual Pooling Module} (Eqs.~\ref{eq:avgpool}, \ref{eq:varpool}); $\mathcal{A}$ is the dual-attention Transformer encoder defined by Eqs.~(\ref{eq:V_temporal})--(\ref{eq:dual_attn_softmax}) and characterised by Proposition~\ref{prop:simplex}; and $\mathcal{H}$ is the convolutional fusion classifier of \hyperref[subsubsec:classification_head]{Classification Head} (Eqs.~\ref{eq:concat}--\ref{eq:logits}).

\subsubsection*{Multi-scale Temporal Convolution}
\label{subsubsec:multiscale_temporal}
EEG signals contain oscillatory components across multiple frequency bands (delta: 0.5--4\,Hz, theta: 4--8\,Hz, alpha: 8--13\,Hz, beta: 13--30\,Hz, low-gamma: 30--50\,Hz), each associated with distinct cognitive processes. To capture these diverse temporal dynamics, a multi-scale temporal convolution module $\mathcal{T}$ comprising four parallel branches with kernel widths $k_i \in \{32, 64, 96, 128\}$ samples is applied to the input.

Given $\mathbf{X} \in \mathbb{R}^{B \times C \times T}$, the output of the $i$-th branch is
\begin{equation}
\mathbf{X}_i = \mathrm{Conv}^{\text{temp}}_{1 \times k_i}(\mathbf{X}), \quad i \in \{1,2,3,4\},
\label{eq:Xi}
\end{equation}
where each branch produces $F_t/4$ output channels at the same temporal length $T$ via reflection padding. At a sampling rate of $256$\,Hz the four widths correspond to approximately $125$, $250$, $375$, and $500$\,ms, covering the periods of the principal EEG rhythms. The branch outputs are concatenated along the filter dimension to produce
\begin{equation}
\mathbf{X}_{\text{temp}} = \mathrm{Concat}(\mathbf{X}_1, \mathbf{X}_2, \mathbf{X}_3, \mathbf{X}_4) \in \mathbb{R}^{B \times F_t \times C \times T}.
\label{eq:Xtemp}
\end{equation}

\begin{remark}[Receptive field]
\label{rem:receptive_field}
The maximum effective receptive field of $\mathcal{T}$ in the temporal direction equals $k_{\max} = 128$ samples ($\approx 500\,$ms at $256$\,Hz). Subsequent operators inherit and extend this support: $\mathcal{S}$ does not alter temporal extent, the average-pooling stride $s$ collapses it to length $L = \lfloor (T-k)/s\rfloor + 1$, and $\mathcal{A}$ has unbounded receptive field (every output position attends to every input position via $\mathrm{softmax}(\mathbf{Q}\mathbf{K}^{\top}/\sqrt{d_k})$). The architecture therefore combines explicit local multi-scale support up to $\approx 500\,$ms with implicit global integration across the full epoch. This combination is necessary for inner-speech decoding because discriminative oscillatory patterns evolve on both fast (theta-locked phase, $\sim$150\,ms) and slow (alpha-band ERD envelope, $\sim$500--1000\,ms) timescales~\cite{nieto2022thinking,lotte2018review,imaginedspeech_review2025}.
\end{remark}

\subsubsection*{Spatial Convolution}
\label{subsubsec:spatial_conv}

Following temporal feature extraction, a depthwise spatial convolution $\mathcal{S}$ is applied to capture the inter-channel dependencies across EEG electrodes:
\begin{equation}
\mathbf{X}_{\text{spat}} = \mathrm{ELU}\bigl(\mathrm{BN}(\mathrm{Conv2D}_{C \times 1}(\mathbf{X}_{\text{temp}}))\bigr) \in \mathbb{R}^{B \times F \times 1 \times T},
\label{eq:Xspat}
\end{equation}

where \(\text{Conv2D}_{C \times 1}\) denotes a convolutional layer with a kernel spanning all \(C\) channels and a single temporal step, \(\text{BN}\) represents batch normalization to stabilize learning and reduce internal covariate shift, and Exponential Linear Unit (ELU) introduces non-linearity while maintaining smooth gradients for negative inputs.

The $C\times1$ kernel jointly weights every electrode at each temporal feature, allowing the model to learn spatial filters tuned to synchronisation and desynchronisation patterns across cortical regions; the resulting feature map encodes both the multi-scale temporal dynamics of $\mathcal{T}$ and the inter-channel spatial structure on which inner-speech decoding depends~\cite{lotte2018review,nieto2022thinking}.
\subsubsection*{Dual Pooling Module}\phantomsection \label{subsec:dual_pooling}

The dual pooling module extracts complementary first- and second-order statistical descriptors from the spatially convolved feature map. Standard average pooling captures the central tendency of the signal but is invariant to the band-limited power modulation that carries event-related (de)synchronisation~\cite{craik2019deep, roy2019chronnectome, vafaei2025transformers}, the dominant discriminative cue for inner-speech EEG (Lemma~\ref{lem:logvar_psd}). VC-DATNet therefore pairs the mean stream with a parallel log-variance stream and routes both through independent invocations of the dual-attention encoder.

\paragraph{Average Pooling.} Given an input feature map \(\mathbf{Z} \in \mathbb{R}^{C \times T}\), the average pooling operation computes the mean over each non-overlapping temporal window of size \(k\) with stride \(s\):

\begin{equation}
\mathbf{P}_{\text{avg}}(j) = \frac{1}{k} \sum_{i=0}^{k-1} \mathbf{Z}(:,\, j \cdot s + i), \quad j = 0, \dots, L-1,
\label{eq:avgpool}
\end{equation}

where \(L = \left\lfloor \frac{T - k}{s} \right\rfloor + 1\) denotes the number of pooling steps.

\paragraph{Variance Pooling.} To capture the local signal variability, variance pooling is defined as:

\begin{equation}
\mathbf{P}_{\text{var}}(j) = \log \left( \max \left( \min \left( \text{Var} \left( \mathbf{Z}(:,\, j \cdot s : j \cdot s + k - 1) \right), 10^6 \right), 10^{-6} \right) \right),
\label{eq:varpool}
\end{equation}

where \(\text{Var}(\cdot)\) computes the element-wise variance across the temporal dimension, and \(\log\) applies a logarithmic transformation to stabilize training and compress the dynamic range. The clamping via \(\min\) and \(\max\) prevents extreme variance values that could lead to numerical instability during optimization.

This dual pooling mechanism retains both the average level of activity and the signal's local temporal variability. The link between the log-variance feature and the physiological substrate of inner-speech decoding, namely event-related (de)synchronisation in narrow frequency bands, is made precise by the following lemma.

\begin{lemma}[Log-variance pooling estimates band-limited log-power]
\label{lem:logvar_psd}
Let $z(t)$ be the output of one of the $\mathcal{T}$ branches with kernel $k_i$, acting as a finite-impulse-response band-pass filter with effective passband $\mathcal{B}_i \subset \mathbb{R}$ around centre frequency $f_i$, and let $\widehat{P}_j := \mathrm{Var}_{\mathcal{W}_j}(z)$ denote the sample variance of $z$ over a temporal window $\mathcal{W}_j$ of length $k$ (the quantity that, after clamping, is log-transformed in Eq.~\ref{eq:varpool}). For a zero-mean wide-sense stationary signal $z(t)$,
\begin{equation}
\mathbb{E}\!\left[\widehat{P}_j\right] \;=\; \int_{\mathcal{B}_i} S_z(f)\,\mathrm{d}f \;=:\; P_{\mathcal{B}_i}(z),
\label{eq:psd_link}
\end{equation}
where $S_z(f)$ is the power spectral density of $z$ and $P_{\mathcal{B}_i}(z)$ its band-limited power. Consequently the pooled value $\mathbf{P}_{\sigma^2}(j) = \log\widehat{P}_j$ satisfies
\begin{equation}
\mathbb{E}\!\left[\mathbf{P}_{\sigma^2}(j)\right] \;=\; \log P_{\mathcal{B}_i}(z) \;-\; \beta_j, \qquad 0 \le \beta_j \le \frac{\mathrm{Var}(\widehat{P}_j)}{2\,P_{\mathcal{B}_i}(z)^2} + o\!\left(\frac{\mathrm{Var}(\widehat{P}_j)}{P_{\mathcal{B}_i}(z)^2}\right),
\label{eq:logvar_bias}
\end{equation}
i.e.\ $\mathbf{P}_{\sigma^2}(j)$ is a (negatively) biased estimator of $\log P_{\mathcal{B}_i}(z)$ whose bias $\beta_j$ vanishes as the relative variance of the windowed estimator does; in particular $\beta_j = O(1/k_{\mathrm{eff}})$, where $k_{\mathrm{eff}}$ is the number of effectively independent samples in the window. Thus the log-variance pooling stream tracks the logarithm of band-limited power, the quantity modulated by event-related desynchronisation (ERD) and synchronisation (ERS)~\cite{lotte2018review,imaginedspeech_review2025}. The average pooling stream $\mathbf{P}_{\mu}$, by contrast, tracks first-moment amplitude and is invariant under sign flips of zero-mean band-pass output, so it cannot recover ERD/ERS information.
\end{lemma}
\begin{proof}[Sketch]
Equation~(\ref{eq:psd_link}) is the Wiener--Khinchin / Parseval identity: for a zero-mean WSS process the variance equals the integral of the PSD, and for a band-pass FIR output the PSD is supported (up to roll-off) on $\mathcal{B}_i$, so the integral reduces to band-limited power~\cite{stoica2005spectral}. For~(\ref{eq:logvar_bias}), apply a second-order Taylor (delta-method) expansion of $\log(\cdot)$ about $\mathbb{E}[\widehat{P}_j] = P_{\mathcal{B}_i}(z)$: $\mathbb{E}[\log\widehat{P}_j] = \log P_{\mathcal{B}_i}(z) - \tfrac12 \mathrm{Var}(\widehat{P}_j)/P_{\mathcal{B}_i}(z)^2 + o(\cdot)$, with the leading term non-negative; by Jensen's inequality $\mathbb{E}[\log\widehat{P}_j] \le \log P_{\mathcal{B}_i}(z)$ exactly, so $\beta_j \ge 0$. The clamping in Eq.~(\ref{eq:varpool}) bounds $\widehat{P}_j$ away from $0$ and $\infty$, ensuring the expansion is valid; for a window of length $k$ with bandwidth $|\mathcal{B}_i|$ the estimator variance scales as $\mathrm{Var}(\widehat{P}_j) = O(P_{\mathcal{B}_i}(z)^2 / k_{\mathrm{eff}})$ with $k_{\mathrm{eff}} \asymp k\,|\mathcal{B}_i|$, giving $\beta_j = O(1/k_{\mathrm{eff}})$.
\end{proof}

Lemma~\ref{lem:logvar_psd} formalises why a parallel mean-and-variance stream is a natural primitive for inner-speech EEG: the variance stream provides a per-window proxy for the spectral signature of ERD/ERS, while the mean stream retains complementary first-moment information that variance alone discards. The recurrence of variance pooling as an independent design choice in MI-specialised architectures~\cite{han2024edpnet} is consistent with this account; the contribution of VC-DATNet is to couple this stream with the dual-attention encoder defined next, rather than to feed it directly to a classifier.

\subsubsection*{Dual Attention Transformer Encoder}\phantomsection \label{subsec:dual_attention}
The dual-attention Transformer encoder (panel~(d) of Figure~\ref{fig:overview}) sits in the same architectural family as recent dual-attention EEG decoders. Three concrete properties define its operational form, applied uniformly across all three datasets: (i) the temporal and spatial gates act on the value tensor $\mathbf{V}$ \emph{before} the standard $\mathrm{softmax}(\mathbf{Q}\mathbf{K}^{\top}/\sqrt{d_k})$ aggregation, leaving the query--key similarity structure intact while suppressing the content routed through it; (ii) the gates are derived asymmetrically (the temporal gate from $\mathbf{V}$ at full sequence resolution, the spatial gate from the sequence-mean summary $\bar{\mathbf{V}}$), assigning different temporal resolutions to the millisecond-scale dynamics of $\mathbf{V}$ and the trial-stationary informativeness of each feature channel; (iii) the encoder is instantiated twice in parallel, one stream over the mean pool and one over the log-variance pool. Lemma~\ref{lem:logvar_psd} and Proposition~\ref{prop:simplex} below formalise the consequences of these choices; the components themselves are recombinations of existing primitives, and we make no primacy claim for any of them in isolation.

The transformer encoder begins by projecting the input features into query ($\mathbf{Q}$), key ($\mathbf{K}$), and value ($\mathbf{V}$) representations through learnable linear transformations:

\begin{equation}
\begin{aligned}
\mathbf{Q} &= \mathbf{W}_Q\mathbf{X} \\
\mathbf{K} &= \mathbf{W}_K\mathbf{X} \\
\mathbf{V} &= \mathbf{W}_V\mathbf{X}
\end{aligned}
\end{equation}

where $\mathbf{W}_Q, \mathbf{W}_K, \mathbf{W}_V \in \mathbb{R}^{d_{\text{model}} \times d_k}$ are learnable parameter matrices, and $d_{\text{model}}$ and $d_k$ denote the model dimension and key dimension, respectively.

\FloatBarrier

The dual attention mechanism consists of two complementary attention pathways operating on the value tensor: a temporal pathway that produces a per-timestep gate from $\mathbf{V}$ at full sequence resolution, and a spatial pathway that produces a per-channel gate from the sequence-mean of $\mathbf{V}$. The asymmetric derivation is deliberate: temporal selectivity must respect the millisecond-scale dynamics of event-related desynchronisation, whereas channel informativeness is comparatively stationary across a trial and benefits from a sequence-level summary. The temporal pathway is formulated as:

\begin{equation}
\mathbf{V}_{\text{temporal}} = \mathbf{V}_{\text{reshaped}} \odot \sigma\left(\mathbf{W}_2 \cdot \text{ReLU}(\mathbf{W}_1 \cdot \mathbf{V}_{\text{reshaped}})\right)
\label{eq:V_temporal}
\end{equation}

where $\mathbf{V}_{\text{reshaped}} \in \mathbb{R}^{B \times N \times (H \cdot D)}$ is the value tensor reshaped to apply attention across the sequence dimension ($N$ represents the sequence length, $H$ is the number of attention heads, and $D$ is the dimension per head), $\sigma$ denotes the sigmoid activation function that transforms the attention weights to the range $[0,1]$, $\mathbf{W}_1 \in \mathbb{R}^{(H \cdot D) \times \frac{(H \cdot D)}{2}}$ and $\mathbf{W}_2 \in \mathbb{R}^{\frac{(H \cdot D)}{2} \times (H \cdot D)}$ are learnable parameters of the temporal attention module, ReLU is the rectified linear unit activation function, and $\odot$ represents element-wise multiplication. The operation assigns a per-timestep weight to each segment of the value tensor, focusing capacity on diagnostically relevant events and attenuating uninformative segments.

The spatial attention pathway complements the temporal one by weighting feature channels by their global informativeness, derived from the sequence-mean of $\mathbf{V}$:

\begin{equation}
\mathbf{V}_{\text{spatial}} = \mathbf{V} \odot \sigma\left(\mathbf{W}_4 \cdot \text{ReLU}(\mathbf{W}_3 \cdot \bar{\mathbf{V}})\right)
\label{eq:V_spatial}
\end{equation}

where $\bar{\mathbf{V}} \in \mathbb{R}^{B \times 1 \times d_{\text{model}}}$ is the sequence-mean of $\mathbf{V}$, providing a per-channel global summary, and $\mathbf{W}_3 \in \mathbb{R}^{d_{\text{model}} \times \frac{d_{\text{model}}}{2}}$ and $\mathbf{W}_4 \in \mathbb{R}^{\frac{d_{\text{model}}}{2} \times d_{\text{model}}}$ are the learnable parameters of the spatial gate. The gate routes feature-channel capacity toward channels with task-relevant content and attenuates the rest.

The two pathways are summed to form the gated value tensor:

\begin{equation}
\mathbf{V}_{\text{dual}} = \mathbf{V}_{\text{temporal}} + \mathbf{V}_{\text{spatial}}
\label{eq:V_dual}
\end{equation}

This dual-attentive value tensor is then used in the standard attention operation:

\begin{equation}
\text{Attention}(\mathbf{Q}, \mathbf{K}, \mathbf{V}_{\text{dual}}) = \text{softmax}\left(\frac{\mathbf{Q}\mathbf{K}^T}{\sqrt{d_k}}\right)\mathbf{V}_{\text{dual}}
\label{eq:dual_attn_softmax}
\end{equation}

where the softmax normalises the attention scores along the key dimension and the scaling factor $\sqrt{d_k}$ stabilises the gradients during backpropagation.

The choice to gate $\mathbf{V}$ \emph{before} the softmax, rather than gating the rows of the attention matrix or the output of the attention block, has a structural consequence that distinguishes VC-DATNet from CBAM-style and post-attention dual-attention designs (\hyperref[sec:related]{Related Work}). The following proposition makes this consequence precise.

\begin{proposition}[Pre-softmax $\mathbf{V}$-gating preserves the attention probability simplex]
\label{prop:simplex}
Let $\mathbf{A} = \mathrm{softmax}(\mathbf{Q}\mathbf{K}^{\top}/\sqrt{d_k}) \in [0,1]^{N\times N}$, so that each row $\mathbf{A}_{i,:}$ lies in the probability simplex $\Delta^{N-1}$, and let $\mathbf{V}, \mathbf{V}_{\mathrm{dual}} \in \mathbb{R}^{N\times D}$. For any element-wise gate $\mathbf{V}_{\mathrm{dual}} = \mathbf{V} \odot \mathbf{G}$ with $\mathbf{G} \in [0,1]^{N\times D}$, the pre-softmax-gated output
\begin{equation}
\mathbf{Y}_{\mathrm{pre}} = \mathbf{A}\,\mathbf{V}_{\mathrm{dual}} = \mathbf{A}\,(\mathbf{V} \odot \mathbf{G})
\label{eq:Ypre}
\end{equation}
satisfies
\begin{equation}
\mathbf{Y}_{\mathrm{pre}}[i,d] \in \mathrm{conv}\bigl(\{\mathbf{V}[j,d] \cdot \mathbf{G}[j,d]\}_{j=1}^{N}\bigr) \quad \forall (i,d),
\label{eq:Ypre_conv}
\end{equation}
i.e.\ each entry of $\mathbf{Y}_{\mathrm{pre}}$ is a convex combination of gated value entries. By contrast, the post-softmax-gated output
\begin{equation}
\mathbf{Y}_{\mathrm{post}} = (\mathbf{A} \odot \mathbf{G}_{\mathrm{post}})\,\mathbf{V}, \qquad \mathbf{G}_{\mathrm{post}} \in [0,1]^{N\times N}
\label{eq:Ypost}
\end{equation}
breaks the simplex structure: in general $\sum_{j} (\mathbf{A} \odot \mathbf{G}_{\mathrm{post}})[i,j] \neq 1$, so $\mathbf{Y}_{\mathrm{post}}[i,:]$ is not a convex combination of value rows.
\end{proposition}
\begin{proof}
For (\ref{eq:Ypre_conv}), fix $(i,d)$. Then $\mathbf{Y}_{\mathrm{pre}}[i,d] = \sum_{j=1}^N \mathbf{A}[i,j]\,(\mathbf{V}[j,d]\cdot \mathbf{G}[j,d])$ with $\mathbf{A}[i,j] \geq 0$ and $\sum_j \mathbf{A}[i,j] = 1$ (softmax property), so $\mathbf{Y}_{\mathrm{pre}}[i,d]$ is by definition a convex combination of $\{\mathbf{V}[j,d]\cdot \mathbf{G}[j,d]\}_j$. For the post-softmax case, write $\mathbf{B} = \mathbf{A} \odot \mathbf{G}_{\mathrm{post}}$. Each row sum is $\sum_j \mathbf{B}[i,j] = \sum_j \mathbf{A}[i,j]\,\mathbf{G}_{\mathrm{post}}[i,j] \leq \sum_j \mathbf{A}[i,j] = 1$, with equality iff $\mathbf{G}_{\mathrm{post}}[i,j]=1$ for every $j$ in the support of $\mathbf{A}[i,:]$. Whenever any gate value is strictly less than 1 on a positive-mass attention entry, the row sum is strictly less than 1 and $\mathbf{B}[i,:]$ is not a probability vector; consequently $\mathbf{Y}_{\mathrm{post}}[i,:]$ is not a convex combination of value rows.
\end{proof}

\begin{remark}[Implication of Proposition~\ref{prop:simplex}]
\label{rem:gate_implication}
Proposition~\ref{prop:simplex} formalises the structural difference between pre- and post-softmax gating. Pre-softmax gating (VC-DATNet) preserves the convex-combination interpretation of attention, so that every output entry remains a probability-weighted average of gated value entries, and therefore inherits the well-conditioned numerical behaviour of softmax attention. Post-softmax gating, used in CBAM-derived dual-attention designs that suppress entries of the attention matrix~\cite{du2022dual,liu2025dmsacnn}, breaks this property: the row sums fall below one, and the gated mixture is no longer interpretable as a convex combination of value vectors. The practical consequence is that gradients of $\mathbf{Y}_{\mathrm{post}}$ with respect to $\mathbf{Q}, \mathbf{K}$ couple non-trivially to the gate values, complicating optimisation; for $\mathbf{Y}_{\mathrm{pre}}$ the factor $\mathbf{A}=\mathrm{softmax}(\mathbf{Q}\mathbf{K}^{\top}/\sqrt{d_k})$ is computed from $\mathbf{Q}$ and $\mathbf{K}$ alone and does not depend on the gate $\mathbf{G}$, so the $\mathbf{Q},\mathbf{K}$ pathway and the gate pathway contribute additively to $\partial\mathbf{Y}_{\mathrm{pre}}$ and do not interact in the backward pass.
\end{remark}

\begin{remark}[Computational cost of the dual-attention block]
\label{rem:complexity}
For a single layer, scaled dot-product attention with $H$ heads of dimension $d_k = D/H$ over a token sequence of length $L$ requires $\Theta(L^2 D + L D^2)$ operations and $4D^2 + 4D$ parameters for the $\mathbf{Q}/\mathbf{K}/\mathbf{V}/\mathrm{out}$ projections. The dual-attention block adds two factorised MLPs: the temporal gate (Eq.~\ref{eq:V_temporal}) contributes $D^2/2 + D^2/2 = D^2$ parameters and $\Theta(L D^2)$ operations; the spatial gate (Eq.~\ref{eq:V_spatial}) contributes the same parameter count but only $\Theta(D^2)$ operations because the MLP is applied to the sequence-mean $\bar{\mathbf{V}} \in \mathbb{R}^{B\times 1 \times D}$. The combined dual-attention overhead is therefore $2D^2$ parameters per layer; with $D=32$ and depth $4$ this is of order $8\times10^3$ parameters in total, a small additive term relative to the model's parameter budget, and it remains asymptotically dominated by the $L^2 D$ similarity computation in $\mathbf{Q}\mathbf{K}^{\top}$ for any $L \geq D$ (here $L=64$, $D=32$).
\end{remark}

Three properties of this construction warrant explicit comparison with prior dual-attention EEG architectures (\hyperref[sec:related]{Related Work}). First, the gating in Eqs.~(\ref{eq:V_temporal})--(\ref{eq:V_dual}) acts on $\mathbf{V}$ \emph{before} the softmax aggregation in Eq.~(\ref{eq:dual_attn_softmax}), so that the $\mathbf{Q}\mathbf{K}^{\top}/\sqrt{d_k}$ similarity structure is preserved intact while the content routed through that structure is suppressed. CBAM-derived dual-attention designs~\cite{du2022dual,liu2025dmsacnn,chen2024damgcn} place their gating outside the attention block (most commonly without one), and decoupled-axis Transformer designs~\cite{pan-2024} apply attention at the embedding or residual level rather than at $\mathbf{V}$. Second, the asymmetric derivation of the two gates (a per-timestep MLP for the temporal axis, a sequence-mean MLP for the spatial axis) encodes a directional inductive bias that symmetric CBAM-style gating and parallel single-attention towers~\cite{li2024dual,sstaf2025} do not provide. Third, because the encoder is instantiated twice (once per statistical stream, \hyperref[subsec:dual_pooling]{Dual Pooling Module}), the spatial gate of the variance pathway can localise discriminative content even when amplitudes are too low for the mean pathway to distinguish, a property of direct relevance for inner speech, where the discriminative signal lives almost entirely in second-order statistics. EDPNet's multi-scale variance pooling~\cite{han2024edpnet} provides the closest variance-pooling antecedent but feeds variance directly to the classifier rather than using it to condition Transformer attention. The factorised gating adds only $\Theta(D^2)$ parameters per gate, below a tenth of the model's parameter budget in total (Remark~\ref{rem:complexity}), and each gate produces an independent attention map that can be inspected to interpret which time windows and electrodes drove a given prediction.

The complete transformer encoder layer integrates this dual attention mechanism with a position-wise feed-forward network:

\begin{equation}
\begin{aligned}
\mathbf{X}' &= \mathbf{X} + \text{DualAttention}(\text{LayerNorm}(\mathbf{X})) \\
\mathbf{X}'' &= \mathbf{X}' + \text{FFN}(\text{LayerNorm}(\mathbf{X}'))
\end{aligned}
\end{equation}

where LayerNorm denotes layer normalization, which stabilizes the hidden state dynamics, and FFN is a two-layer feed-forward network with GELU (Gaussian Error Linear Unit) activation:

\begin{equation}
\text{FFN}(\mathbf{X}) = \mathbf{W}_6 \cdot \text{GELU}(\mathbf{W}_5 \cdot \mathbf{X})
\end{equation}

where $\mathbf{W}_5 \in \mathbb{R}^{d_{\text{model}} \times d_{\text{ff}}}$ and $\mathbf{W}_6 \in \mathbb{R}^{d_{\text{ff}} \times d_{\text{model}}}$ are the parameters of the feed-forward network, and $d_{\text{ff}}$ represents its inner dimension.

Stacking four such encoder layers within each statistical stream allows the dual-attention mechanism to refine the value tensor through successive rounds of variance-conditioned gating, while preserving the standard residual+LayerNorm structure that stabilises gradients in small-cohort EEG training~\cite{zhao2024ctnet,ma2024attention}.

\subsubsection*{Classification Head}
\label{subsubsec:classification_head}

The classifier head $\mathcal{H}$ takes the two stream outputs $\mathbf{X}_{\text{avg}}^{\text{transformed}}$ and $\mathbf{X}_{\text{var}}^{\text{transformed}}$ produced by independent invocations of the dual-attention encoder $\mathcal{A}$ on the mean-pooled and log-variance-pooled streams (Definition~\ref{def:datnet}), concatenates them along the feature dimension, applies a single 2D convolutional encoder for local feature integration, and produces class logits via a linear projection:

\begin{equation}
\mathbf{X}_{\text{combined}} = \mathrm{Concat}(\mathbf{X}_{\text{avg}}^{\text{transformed}}, \mathbf{X}_{\text{var}}^{\text{transformed}}),
\label{eq:concat}
\end{equation}
\begin{equation}
\mathbf{X}_{\text{encoded}} = \mathrm{ConvEncoder}(\mathbf{X}_{\text{combined}}).
\end{equation}

Here, $\mathbf{X}_{\text{avg}}^{\text{transformed}}$ and $\mathbf{X}_{\text{var}}^{\text{transformed}}$ denote the outputs of the transformer encoders applied to the average and variance pooling branches, respectively. The convolutional encoder (\texttt{ConvEncoder}) consists of a single 2D convolutional layer, followed by batch normalization and an ELU activation function. This configuration promotes local feature integration across the combined statistical descriptors, reinforcing discriminative patterns for classification \cite{vafaei2025transformers}.

The resulting encoded features are then flattened and passed through a linear layer to generate the classification logits:

\begin{equation}
\mathbf{y} = \mathbf{W}_{\text{classify}} \cdot \mathrm{Flatten}(\mathbf{X}_{\text{encoded}}) \in \mathbb{R}^{B\times K},
\label{eq:logits}
\end{equation}

where $\mathbf{W}_{\text{classify}} \in \mathbb{R}^{d_{\text{flat}} \times K}$ is the linear classifier weight matrix, $d_{\text{flat}}$ the dimensionality of the flattened representation, and $K$ the number of target classes. The model is trained end-to-end by minimising the standard cross-entropy loss
\begin{equation}
\mathcal{L}(\theta) = -\frac{1}{B} \sum_{b=1}^{B} \sum_{c=1}^{K} y^{*}_{b,c}\,\log\,\mathrm{softmax}(\mathbf{y})_{b,c},
\label{eq:ce_loss}
\end{equation}
where $y^{*}$ is the one-hot ground-truth label tensor.

\section*{Results}

This section reports the decoding results. We first state the implementation and the complete hyperparameter configuration (\hyperref[subsec:impl]{Implementation and Hyperparameters}), then report the inner-speech results across the three datasets, the sensor/source two-regime effect, and the cross-method ceiling analysis.

\subsection*{Implementation and Hyperparameters}\phantomsection \label{subsec:impl}

All experiments were implemented in PyTorch and run on a single consumer-grade GPU. Each dataset is decoded under within-subject five-fold stratified cross-validation: for every subject the trials are partitioned into five class-balanced folds, the model is trained on four and evaluated on the held-out fold, and the reported per-subject accuracy is the mean over the five folds. The random seed is fixed at $0$; a multi-seed analysis with confidence intervals is identified as future work (\hyperref[subsec:limitations]{Limitations and Scope}). Following the established convention for this benchmark family, the accuracy of a fold is its best test accuracy over the training run. Each EEG epoch is band-pass filtered to $4$--$40$\,Hz and per-trial $z$-scored before being passed to the model. We decode each dataset from two input representations: the sensor-space EEG, and a cortical source-space projection obtained with an sLORETA inverse solution and averaged within $20$ language-network regions of interest of the Desikan--Killiany atlas (\hyperref[subsec2]{Data Preprocessing}).

One architecture and one training recipe are used for every dataset and both representations; the complete configuration is given in Table~\ref{tab:hyperparams}. The only quantities that change between runs are the two input dimensions the data itself fixes: the number of input channels (sensor-space) or regions of interest (source-space), and the number of output classes. Neither of these is tuned.

\begin{table}[!htbp]
\centering
\footnotesize
\caption{Complete VC-DATNet configuration, identical across all three datasets and both signal representations. The values are the defaults of the released training script. The input dimension (sensor-channel or source-ROI count) and the output-class count are fixed by the dataset and the task; every other quantity is held constant.}
\label{tab:hyperparams}
\begin{tabular}{l|c}
\toprule
\textbf{Component / hyperparameter} & \textbf{Value} \\
\midrule
\multicolumn{2}{@{}l}{\textit{Preprocessing}} \\
\midrule
Sampling rate                         & $256$\,Hz \\
Epoch length                          & $2.5$\,s ($\approx$640 samples) \\
Band-pass filter                      & $4$--$40$\,Hz (4th-order Butterworth) \\
Normalisation                         & per-trial $z$-score \\
\midrule
\multicolumn{2}{@{}l}{\textit{Architecture}} \\
\midrule
Temporal conv.\ kernel widths          & $\{32, 64, 96, 128\}$ samples \\
Filters per temporal branch            & $8$ (concatenated $\to$ embedding dim.\ $32$) \\
Spatial convolution                    & depthwise $(C\times1)$, BatchNorm $+$ ELU \\
Statistical pooling                    & mean and log-variance; window $64$ / stride $16$ \\
Transformer encoder                    & two streams, $4$ dual-attention layers each \\
Embedding dimension $D$ / heads $H$     & $32$ / $8$ ($d_k = 4$) \\
FFN hidden dimension                   & $128$ \\
\midrule
\multicolumn{2}{@{}l}{\textit{Regularisation and training}} \\
\midrule
Stream / attention / FFN dropout        & $0.5$ / $0.3$ / $0.3$ \\
Optimiser                               & AdamW (lr $5\times10^{-4}$, weight decay $5\times10^{-4}$) \\
Learning-rate schedule                  & cosine annealing \\
Batch size / epochs                     & $16$ / $300$ \\
Loss                                    & cross-entropy, label smoothing $0.1$ \\
Augmentation                            & mixup ($\alpha = 0.3$) \\
\midrule
\multicolumn{2}{@{}l}{\textit{Evaluation}} \\
\midrule
Cross-validation                        & within-subject, $5$-fold stratified \\
Random seed                             & $0$ \\
\midrule
\multicolumn{2}{@{}l}{\textit{Dataset-determined}} \\
\midrule
Input dimension $C$                     & sensor: dataset channel count; source: $20$ ROIs \\
Output classes $K$                      & $2$, $4$, or $8$ (task-determined) \\
\bottomrule
\end{tabular}
\end{table}

The baselines compared against in the following tables are the strongest envelope-consistent peer-reviewed results published on each dataset (cited in place), together with the full ten-architecture cross-method bench reported in \hyperref[subsec:ceiling]{A Signal-Limited Accuracy Ceiling}.

\subsection*{Inner-Speech Decoding Across Three Datasets}
\label{subsec:innerspeech_results}

We evaluate VC-DATNet on three independent inner-speech EEG datasets under one fixed protocol: within-subject five-fold cross-validation, EEG modality only. For each dataset we decode from two signal representations (the sensor-space EEG and a cortical source-space projection obtained by an sLORETA inverse solution restricted to language-network regions of interest; \hyperref[subsec2]{Data Preprocessing}), and report both, together with a per-subject \emph{hybrid} that selects, for each subject, whichever representation decoded that subject better. Table~\ref{tab:innerspeech_summary} summarises the headline accuracies against the strongest envelope-consistent peer-reviewed baseline on each dataset.

\begin{table}[!htbp]
\centering
\footnotesize
\caption{Headline inner-speech decoding accuracy of VC-DATNet on the three datasets, against the strongest envelope-consistent peer-reviewed baseline on each. All VC-DATNet results are within-subject five-fold cross-validation, EEG-only. ``Sensor'' and ``source'' denote the two signal representations (\hyperref[subsec2]{Data Preprocessing}); ``hybrid'' selects, per subject, the better of the two. Chance is $100/K$ for a $K$-class task. SyncSpeech has no prior published decoding result.}
\label{tab:innerspeech_summary}
\begin{tabular}{l|c|c|c|c|l}
\toprule
\textbf{Dataset / task} & \textbf{Classes} & \textbf{Chance (\%)} & \textbf{VC-DATNet (\%)} & \textbf{Strongest prior (\%)} & \textbf{Reference} \\
\midrule
ToL, directional                & 4 & 25.0 & \textbf{39.48 (hybrid)} & 36.1 (BiLSTM)        & Gasparini et al.~\cite{Gasparini2022InnerSpeech} \\
BISpeech, 8-word                 & 8 & 12.5 & 20.60 (sensor) & 14.78 (RF, EEG-only) & Wilson et al.~\cite{wilson2023performance} \\
BISpeech, social vs.\ numerical  & 2 & 50.0 & 59.37 (source) & 62.81 (SVM)          & Wilson et al.~\cite{wilson2022bimodal} \\
BISpeech, social 4-word          & 4 & 25.0 & 38.68 (hybrid) & --- (none EEG-only)  & --- \\
BISpeech, numerical 4-word       & 4 & 25.0 & 36.78 (sensor) & --- (none EEG-only)  & --- \\
SyncSpeech, 8-word               & 8 & 12.5 & 25.53 (hybrid) & --- (none published) & --- \\
\bottomrule
\end{tabular}
\end{table}

\paragraph{Thinking Out Loud.}
On the ToL 4-class directional task (10 subjects, full $128$-channel BioSemi montage, vanilla single-condition training), VC-DATNet reaches $37.58\%$ in sensor space and $35.11\%$ in source space as grand means, and $39.48\%$ for the per-subject sensor/source hybrid. The strongest peer-reviewed baseline on this dataset is the BiLSTM of Gasparini et al.~\cite{Gasparini2022InnerSpeech} at $36.1\%$; earlier baselines are lower, with an EEGNet decoder at $29.7\%$ and the dataset descriptor's classical baseline near chance~\cite{nieto2022thinking}. The sensor-space result already exceeds the strongest prior baseline by $1.5$ percentage points, and the hybrid by $3.4$ percentage points (single-seed; multi-seed significance is identified as future work in \nameref{subsec:limitations}).

\paragraph{Two complementary decoding regimes.}
The grand means conceal a consistent per-subject structure (Table~\ref{tab:tol_persubject}). Sensor-space decoding is the better representation for six subjects (S02, S03, S07, S08, S09, S10) and source-space for the other four (S01, S04, S05, S06); both regimes are present. Source-space decoding helps the subjects whose sensor-space signal is weak or spatially diffuse, by concentrating the focal language-network activity that volume conduction smears across the scalp; sensor-space decoding wins for subjects whose scalp signal is already well localised, where the source projection discards usable information. Because one fixed architecture decodes both representations, this split cannot be an artefact of model selection; it is a property of the signal. The per-subject hybrid ($39.48\%$) is the envelope of the two regimes.

\begin{table}[!htbp]
\centering
\footnotesize
\setlength{\tabcolsep}{4pt}
\caption{Per-subject 4-class accuracy (\%) on Thinking Out Loud (10 subjects, within-subject five-fold CV) under two configurations of VC-DATNet. Block (a) uses the cross-architecture bench protocol (full $128$-channel BioSemi montage, vanilla single-condition training, no cross-condition pretrain), matching every other architecture in Table~\ref{tab:ceiling_full}. Block (b) uses the a priori $22$-channel inner-speech-relevant subset with cross-condition (Pron+Inner+Vis) multi-task pretrain followed by inner-only fine-tune. In both blocks the source-space row uses the same $20$-ROI Desikan--Killiany projection. ``Hybrid'' is the per-subject better of sensor and source within each block. The two-regime effect is present under both configurations (sens/src wins re-balance slightly across configs but both regimes co-exist), and the headline result is robust to either choice of channel subset and pretrain.}
\label{tab:tol_persubject}
\begin{tabular}{l|cccccccccc|c}
\toprule
\textbf{Representation} & \textbf{S01} & \textbf{S02} & \textbf{S03} & \textbf{S04} & \textbf{S05} & \textbf{S06} & \textbf{S07} & \textbf{S08} & \textbf{S09} & \textbf{S10} & \textbf{Mean} \\
\midrule
\multicolumn{12}{@{}l}{\textit{(a) Bench protocol: full $128$-channel BioSemi montage, vanilla single-condition training.}} \\
Sensor-space  & 30.0 & 38.8 & 45.0 & 35.0 & 32.5 & 37.0 & 37.1 & 40.0 & 40.4 & 40.0 & 37.58 \\
Source-space  & 40.0 & 31.2 & 32.8 & 38.8 & 36.7 & 38.0 & 36.2 & 32.0 & 30.0 & 35.4 & 35.11 \\
Hybrid (best) & 40.0 & 38.8 & 45.0 & 38.8 & 36.7 & 38.0 & 37.1 & 40.0 & 40.4 & 40.0 & \textbf{39.48} \\
Winner        & src  & sens & sens & src  & src  & src  & sens & sens & sens & sens & --- \\
\midrule
\multicolumn{12}{@{}l}{\textit{(b) A priori subset: $22$ inner-speech-relevant channels, cross-condition (Pron+Inner+Vis) multi-task pretrain $\to$ inner-only fine-tune.}} \\
Sensor-space  & 44.0 & 36.7 & 37.2 & 30.0 & 31.2 & 31.0 & 32.1 & 32.0 & 34.6 & 35.4 & 34.42 \\
Source-space  & 40.0 & 31.2 & 32.8 & 38.8 & 36.7 & 38.0 & 36.2 & 32.0 & 30.0 & 35.4 & 35.11 \\
Hybrid (best) & 44.0 & 36.7 & 37.2 & 38.8 & 36.7 & 38.0 & 36.2 & 32.0 & 34.6 & 35.4 & \textbf{36.95} \\
Winner        & sens & sens & sens & src  & src  & src  & src  & tie  & sens & tie  & --- \\
\bottomrule
\end{tabular}
\end{table}

\paragraph{BISpeech.}
On the BISpeech corpus (4 subjects, EEG only; Table~\ref{tab:bispeech}) VC-DATNet reaches $20.60\%$ on the 8-word task (chance $12.5\%$), $59.37\%$ on the binary social-versus-numerical task (chance $50\%$), and $37.44\%$ and $36.78\%$ on the 4-word social and numerical subtasks (chance $25\%$). The 8-word result is $5.8$ percentage points above the strongest EEG-only baseline (Wilson et al.\ random forest, $14.78\%$~\cite{wilson2023performance}). The binary result is below Wilson et al.'s $62.81\%$~\cite{wilson2022bimodal} as a grand mean, although the best individual subject reaches $63.6\%$. The sensor/source two-regime effect recurs: source-space is the better representation on the binary task and sensor-space on the finer-grained multi-class tasks, consistent with the source projection being well-conditioned for coarse contrasts but under-determined for 8-way discrimination given only four subjects.

\begin{table}[!htbp]
\centering
\footnotesize
\caption{BISpeech (ds004196) decoding accuracy (\%): 4 subjects, within-subject five-fold CV, EEG-only. ``Hybrid'' is the per-subject better representation. The two-regime effect recurs: source-space wins the binary task, sensor-space the finer-grained multi-class tasks.}
\label{tab:bispeech}
\begin{tabular}{l|c|c|c|c|c}
\toprule
\textbf{Task} & \textbf{Classes} & \textbf{Chance (\%)} & \textbf{Sensor (\%)} & \textbf{Source (\%)} & \textbf{Hybrid (\%)} \\
\midrule
8-word                & 8 & 12.5 & \textbf{20.60} & 17.46 & 20.60 \\
Social vs.\ numerical & 2 & 50.0 & 57.64 & \textbf{59.37} & 59.37 \\
Social, 4-word        & 4 & 25.0 & 37.44 & 35.72 & \textbf{38.68} \\
Numerical, 4-word     & 4 & 25.0 & \textbf{36.78} & 35.67 & 36.78 \\
\bottomrule
\end{tabular}
\end{table}

\paragraph{Per-subject comparison against the full bench on BISpeech.}
Table~\ref{tab:bispeech_persubject} reports the per-subject hybrid accuracy of every benched architecture on BISpeech, so that VC-DATNet's grand-mean lead (Table~\ref{tab:bispeech}) is decomposed and the per-subject ranking can be inspected directly. VC-DATNet attains the outright top hybrid accuracy on 2 of 4 subjects (sub-01 and sub-05), ranks second on sub-02 (within $0.31$ percentage points of the winner), and ranks seventh on sub-03 where MSVTNet is the per-subject leader. The mean rank of VC-DATNet across the four subjects is $2.75$ out of ten competing architectures.

\begin{table}[!htbp]
\centering
\footnotesize
\setlength{\tabcolsep}{4pt}
\caption{Per-subject hybrid accuracy (\%) on BISpeech 8-word task across all benched architectures (4 subjects, within-subject five-fold CV, EEG-only). Hybrid is per-subject best of sensor- and source-space. Bold = top accuracy per subject. Underline = VC-DATNet. ``---'' denotes a configuration not run.}
\label{tab:bispeech_persubject}
\begin{tabular}{l|cccc|c}
\toprule
\textbf{Method} & \textbf{S01} & \textbf{S02} & \textbf{S03} & \textbf{S05} & \textbf{Mean} \\
\midrule
EEGNet~\cite{lawhern2018eegnet}                & 17.85 & 18.18 & 18.82 & 19.69 & 18.64 \\
ShallowConvNet~\cite{schirrmeister2017deep}    & 19.73 & 20.38 & 17.57 & 19.38 & 19.27 \\
DeepConvNet~\cite{schirrmeister2017deep}       & 17.87 & 17.88 & 17.87 & 17.81 & 17.86 \\
EEG-TCNet~\cite{ingolfsson2020eeg}             & 18.80 & 16.29 & 18.50 & 17.81 & 17.85 \\
EEGPT~\cite{wang2024eegpt} (linear probe)      & 21.00 & 18.20 & 17.90 & 19.10 & 19.05 \\
TransNet~\cite{ma2024attention}                & 18.81 & 20.37 & 18.80 & 20.31 & 19.57 \\
CTNet~\cite{zhao2024ctnet}                     & 17.85 & 18.17 & 18.18 & 21.56 & 18.94 \\
MSVTNet~\cite{liu2024msvtnet}                  & 18.49 & \textbf{21.63} & \textbf{19.12} & 18.75 & 19.50 \\
EEGNeX~\cite{chen2024eegnex}                   & 19.43 & 20.07 & 17.87 & 21.25 & 19.66 \\
\midrule
\textbf{VC-DATNet (ours)}                      & \underline{\textbf{21.01}} & \underline{21.32} & \underline{17.87} & \underline{\textbf{22.19}} & \underline{\textbf{20.60}} \\
\midrule
\textit{Per-subject winner} & \textit{VC-DATNet} & \textit{MSVTNet} & \textit{MSVTNet} & \textit{VC-DATNet} & --- \\
\bottomrule
\end{tabular}
\end{table}

\paragraph{SyncSpeech.}
SyncSpeech is decoded here for the first time. Across its three subjects (8-word task, chance $12.5\%$; Table~\ref{tab:ds006033}) VC-DATNet reaches $24.51\%$ in sensor space, $23.02\%$ in source space, and $25.53\%$ for the per-subject hybrid. The sensor/source split appears a third time (source-space wins one subject, sensor-space the other two), replicating the two-regime effect on an independently collected corpus and supporting its interpretation as a property of inner-speech EEG rather than of any single dataset.

\begin{table}[!htbp]
\centering
\footnotesize
\caption{SyncSpeech (ds006033) decoding accuracy (\%): 8-word inner speech, 3 subjects, within-subject five-fold CV, EEG-only. To our knowledge these are the first published decoding results on this corpus.}
\label{tab:ds006033}
\begin{tabular}{l|ccc|c}
\toprule
\textbf{Representation} & \textbf{Sub-01} & \textbf{Sub-02} & \textbf{Sub-03} & \textbf{Mean} \\
\midrule
Sensor-space  & 21.45 & 27.42 & 24.65 & 24.51 \\
Source-space  & 24.52 & 26.33 & 18.22 & 23.02 \\
Hybrid (best) & 24.52 & 27.42 & 24.65 & \textbf{25.53} \\
\bottomrule
\end{tabular}
\end{table}

\subsection*{A Signal-Limited Accuracy Ceiling}
\label{subsec:ceiling}

A natural question is whether the modest absolute accuracies above reflect a limitation of VC-DATNet specifically or a ceiling intrinsic to the data. To separate these, we evaluated ten published EEG decoders spanning recurrent, convolutional, TCN, foundation-model, and CNN--Transformer hybrid families on the same three datasets under the same protocol (Table~\ref{tab:ceiling_full}). All decoders are trained with the within-subject five-fold cross-validation, the same 4--40\,Hz bandpass and per-trial $z$-score preprocessing, the same AdamW + cosine-annealing optimiser with mixup ($\alpha=0.3$) and label smoothing ($0.1$), and 120 training epochs. The only thing that varies between rows is the model. Despite spanning over three orders of magnitude in parameter count and five distinct architectural families, the methods converge to the same narrow accuracy band on every dataset: $32$--$38\%$ on ToL (chance $25\%$) and $17$--$27\%$ on the 8-class tasks (chance $12.5\%$). All ten methods fall within a narrow $\sim$5--6\,pp band on every dataset, with the large pretrained foundation model (EEGPT, $\sim$25\,M parameters, evaluated here under its intended linear-probe protocol) sitting within $\sim$2\,pp of the cluster centre and the smallest TCN baseline trailing the cluster by a few percentage points on the 8-class tasks. This convergence is consistent with a signal-limited decoding ceiling on these public corpora under this preprocessing pipeline (4--40\,Hz bandpass, per-trial $z$-score, $20$-ROI source projection, $22$-channel ToL subset) rather than a capacity- or design-limited one; an alternative preprocessing pipeline could in principle shift the cluster up or down, although the within-pipeline result is robust to architectural choice. The bound is informative about what architectural work on these datasets can be expected to deliver, and we regard it as a substantive finding in its own right.

\begin{table}[!htbp]
\centering
\footnotesize
\setlength{\tabcolsep}{3pt}
\caption{Ten EEG decoders evaluated on the same three inner-speech datasets in both signal representations under the same fixed protocol (within-subject five-fold CV, bandpass 4--40\,Hz, per-trial $z$-score, AdamW + cosine schedule, mixup $\alpha=0.3$, label smoothing $0.1$, 120 epochs, EEG-only). The ``Family'' column gives the architectural family (CNN, TCN = temporal convolutional network, RNN, Foundation = self-supervised pretraining, CNN-T = CNN--Transformer hybrid). The ``Domain'' column gives the task each architecture was originally developed and validated for: MI = motor imagery; IS = inner speech; Multi = multiple EEG tasks. Eight of ten trained-from-scratch baselines were originally validated on motor imagery; only Gasparini's BiLSTM was developed on inner speech. ``Params'' is the trainable-parameter count for the canonical ToL configuration ($C\!=\!128$, $T\!=\!641$, $K\!=\!4$); counts differ slightly with $C$ and $K$ on the other datasets. EEGPT's parameter count is the frozen pretrained encoder (Wang et al.~\cite{wang2024eegpt}). For each (model, dataset) we report grand-mean accuracy in sensor space (``S''), source space (``Sr''), and per-subject hybrid (``H'', better of the two). All values \% accuracy. Despite spanning over three orders of magnitude in parameter count, five architectural families, and three task domains, the methods converge to the same narrow accuracy band on each dataset. ``---'' denotes a configuration not run; EEGPT (linear-probe) and the reproduced Gasparini BiLSTM are sensor-only by design.}
\label{tab:ceiling_full}
\begin{tabular}{l|l|l|r|ccc|ccc|ccc}
\toprule
& & & & \multicolumn{3}{c|}{\textbf{ToL (4c)}} & \multicolumn{3}{c|}{\textbf{BISpeech (8c)}} & \multicolumn{3}{c}{\textbf{SyncSpeech (8c)}} \\
\textbf{Method} & \textbf{Family} & \textbf{Domain} & \textbf{Params} & \textbf{S} & \textbf{Sr} & \textbf{H} & \textbf{S} & \textbf{Sr} & \textbf{H} & \textbf{S} & \textbf{Sr} & \textbf{H} \\
\midrule
EEGNet (Lawhern et al.)~\cite{lawhern2018eegnet}                                & CNN        & Multi       & 4.4\,K   & 29.7$^\dagger$ & --- & 29.7$^\dagger$ & 18.40 & 17.93 & 18.64 & 21.90 & 19.74 & 22.34 \\
ShallowConvNet (Schirrmeister et al.)~\cite{schirrmeister2017deep}              & CNN        & MI          & 211.8\,K & 34.70 & 32.96 & 35.21 & 19.04 & 18.40 & 19.27 & 22.53 & 23.23 & 26.41 \\
DeepConvNet (Schirrmeister et al.)~\cite{schirrmeister2017deep}                 & CNN        & MI          & 345.9\,K & 35.57 & 31.62 & 35.57 & 17.47 & 16.84 & 17.86 & 21.47 & 23.69 & 23.70 \\
EEGNeX (Chen et al.)~\cite{chen2024eegnex}                                      & CNN        & Multi       & 62.9\,K  & 35.08 & 35.44 & 36.68 & 19.10 & 18.95 & 19.66 & 26.75 & 24.20 & 26.75 \\
EEG-TCNet (Ingolfsson et al.)~\cite{ingolfsson2020eeg}                          & TCN        & MI          & 5.9\,K   & 31.28 & 30.34 & 31.99 & 17.70 & 15.19 & 17.85 & 15.23 & 14.61 & 16.81 \\
Gasparini et al.\ BiLSTM~\cite{Gasparini2022InnerSpeech} (reprod.)              & RNN        & \textbf{IS} & 305.6\,K & 36.19 & --- & 36.19 & --- & --- & --- & --- & --- & --- \\
EEGPT (Wang et al.)~\cite{wang2024eegpt} (linear probe)                         & Foundation & Multi       & 25.0\,M  & 35.50 & --- & 35.50 & 19.05 & --- & 19.05 & 23.27 & --- & 23.27 \\
TransNet (Ma et al.)~\cite{ma2024attention}                                     & CNN-T      & MI          & 191.8\,K & 36.13 & 35.28 & 36.79 & 19.26 & 18.87 & 19.57 & 24.71 & 21.52 & 24.73 \\
CTNet (Zhao et al.)~\cite{zhao2024ctnet}                                        & CNN-T      & MI          & 156.1\,K & 33.37 & 33.06 & 34.47 & 18.94 & 17.30 & 19.02 & 21.92 & 19.74 & 22.21 \\
MSVTNet (Liu et al.)~\cite{liu2024msvtnet}                                      & CNN-T      & MI          & 81.0\,K  & 34.02 & 31.82 & 34.14 & 18.95 & 18.56 & 19.50 & 23.36 & 22.81 & 24.17 \\
\midrule
\textbf{VC-DATNet (ours)}                                  & CNN-T      & \textbf{IS} & \textbf{131.3\,K} & 34.42 & 35.11 & \textbf{36.95} & \textbf{20.60} & 17.46 & \textbf{20.60} & 24.51 & 23.02 & 25.53 \\
\textbf{VC-DATNet (ours)}$^{\ddagger}$                     & CNN-T      & \textbf{IS} & \textbf{131.3\,K} & \textbf{37.58} & 35.11 & \textbf{39.48} & --- & --- & --- & --- & --- & --- \\
\midrule
Chance                                                     & ---        & ---         & ---      & 25.0 & 25.0 & 25.0 & 12.5 & 12.5 & 12.5 & 12.5 & 12.5 & 12.5 \\
\bottomrule
\end{tabular}
\smallskip

{\footnotesize $^{\dagger}$ Reported value from van den Berg et al.~\cite{vandenberg2021inner}, an EEGNet-style decoder on 8 ToL subjects; not re-run by us under our pipeline on this dataset.\\
$^{\ddagger}$ Second VC-DATNet row reports a re-run on the same 10 ToL subjects using the cross-architecture bench's standard sensor protocol (full $128$-channel BioSemi montage, vanilla single-condition training, no cross-condition pretrain), matching every other row of this table. The first VC-DATNet row uses VC-DATNet's headline protocol (a priori $22$-channel inner-speech-relevant subset with cross-condition Pron+Inner+Vis multi-task pretrain $\to$ inner-only fine-tune; see Table~\ref{tab:tol_persubject}). Both rows share the same source-space configuration ($20$-ROI Desikan--Killiany projection) so the Sr value is identical; the H column is the per-subject best of the two representations within each protocol. BISpeech and SyncSpeech columns are not duplicated because VC-DATNet uses a single protocol on those corpora.}
\end{table}

\paragraph{Cross-architecture replication of the two-regime effect.}
A central claim of this paper is that the sensor/source split reported for VC-DATNet (\hyperref[subsec:innerspeech_results]{Inner-Speech Decoding Across Three Datasets}) is a property of the inner-speech signal rather than of the decoder used. The full cross-architecture bench provides an independent test of this claim: if the split is a data property, it must recur across architecturally distinct models. Table~\ref{tab:tworegime_replication} tabulates, for each (model, dataset) cell of Table~\ref{tab:ceiling_full}, the number of subjects on which sensor-space decoding outperformed source-space, the number on which source outperformed sensor, and the number of ties (within 0.5\,pp). With one exception (DeepConvNet on ToL, which prefers sensor for all 10 subjects), every model on every dataset shows both regimes: at least one subject decoded better from sensor space and at least one from source space. This recurs across recurrent, convolutional, TCN, and CNN--Transformer hybrid families, on three independently collected corpora. The two-regime effect is therefore not an artefact of VC-DATNet's architecture; it is a property of how inner-speech-related cortical signal projects onto scalp EEG.

\begin{table}[!htbp]
\centering
\footnotesize
\setlength{\tabcolsep}{4pt}
\caption{Cross-architecture replication of the sensor/source two-regime effect. For each (model, dataset) cell of Table~\ref{tab:ceiling_full}, we count the number of subjects on which sensor-space decoding beat source-space (``Sens$\uparrow$''), source-space beat sensor-space (``Src$\uparrow$''), or the two tied to within 0.5\,pp (``Tie''). The total per cell is the number of subjects in that dataset (ToL: 10, BISpeech: 4, SyncSpeech: 3). Both regimes appear in every (model, dataset) cell except DeepConvNet on ToL, confirming the split is a data property and not specific to VC-DATNet.}
\label{tab:tworegime_replication}
\begin{tabular}{l|ccc|ccc|ccc}
\toprule
& \multicolumn{3}{c|}{\textbf{ToL ($n\!=\!10$)}} & \multicolumn{3}{c|}{\textbf{BISpeech ($n\!=\!4$)}} & \multicolumn{3}{c}{\textbf{SyncSpeech ($n\!=\!3$)}} \\
\textbf{Method} & \textbf{Sens$\uparrow$} & \textbf{Src$\uparrow$} & \textbf{Tie} & \textbf{Sens$\uparrow$} & \textbf{Src$\uparrow$} & \textbf{Tie} & \textbf{Sens$\uparrow$} & \textbf{Src$\uparrow$} & \textbf{Tie} \\
\midrule
ShallowConvNet  & 7 & 2 & 1 & 1 & 1 & 2 & 1 & 2 & 0 \\
DeepConvNet     & 10 & 0 & 0 & 3 & 1 & 0 & 0 & 2 & 1 \\
EEG-TCNet       & 7 & 2 & 1 & 3 & 1 & 0 & 2 & 1 & 0 \\
EEGNet          & --- & --- & --- & 3 & 1 & 0 & 2 & 1 & 0 \\
TransNet        & 6 & 4 & 0 & 1 & 2 & 1 & 2 & 0 & 1 \\
CTNet           & 6 & 4 & 0 & 2 & 0 & 2 & 2 & 1 & 0 \\
MSVTNet         & 6 & 1 & 3 & 2 & 1 & 1 & 1 & 2 & 0 \\
EEGNeX          & 5 & 4 & 1 & 1 & 1 & 2 & 3 & 0 & 0 \\
\textbf{VC-DATNet} & 4 & 4 & 2 & 4 & 0 & 0 & 2 & 1 & 0 \\
\bottomrule
\end{tabular}
\end{table}

\subsection*{Class-Conditional Cortical Signature of Inner Speech}
\label{subsec:cortical_signature}

\begin{figure}[H]
  \centering
  \begin{subfigure}[t]{0.78\linewidth}
    \centering
    \includegraphics[width=\linewidth]{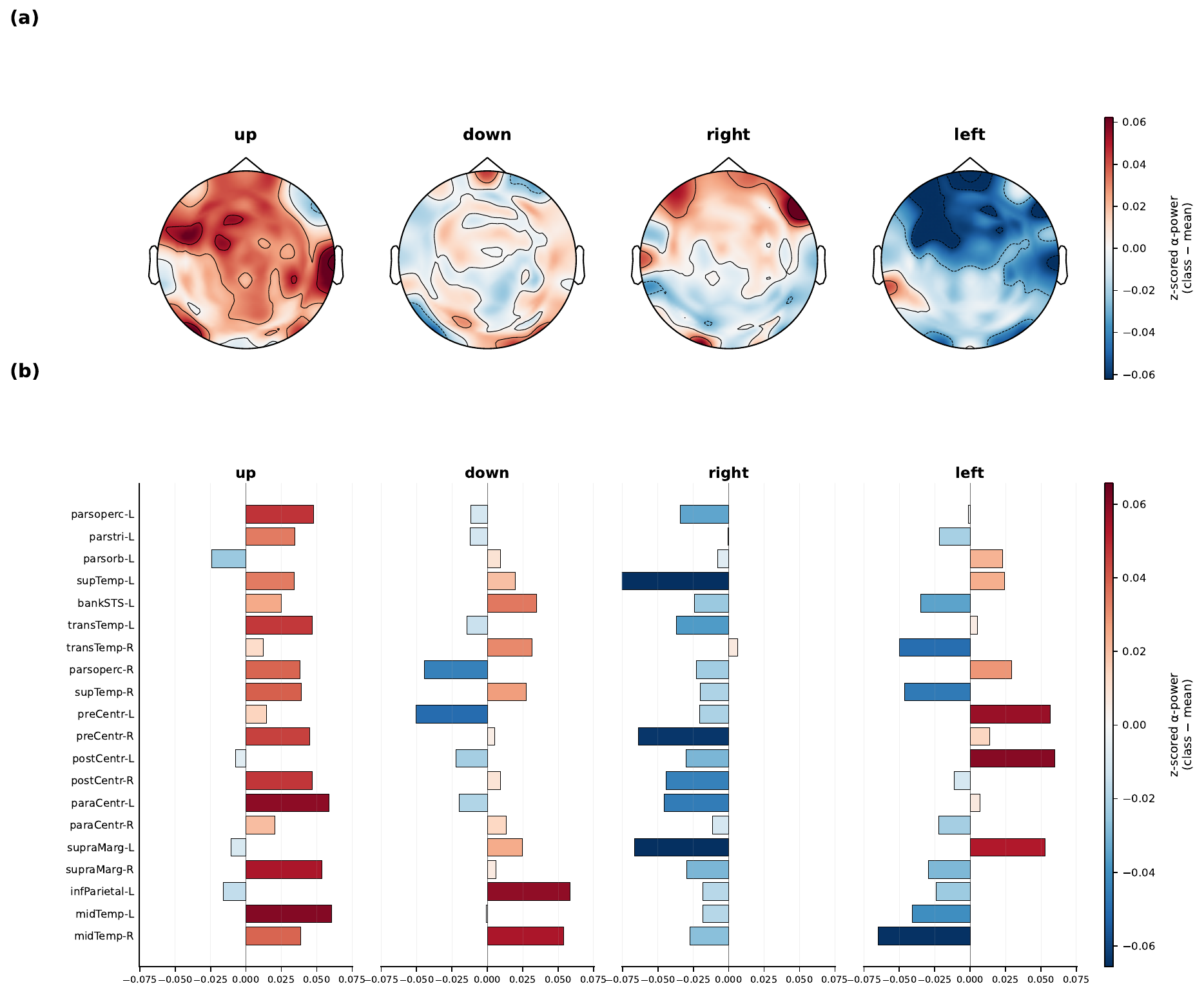}
    \caption{Thinking Out Loud (ds003626, $10$ subjects, $4$ directional classes: \textit{up}, \textit{down}, \textit{right}, \textit{left}).}
    \label{fig:story_tol}
  \end{subfigure}\\[2pt]
  \begin{subfigure}[t]{0.78\linewidth}
    \centering
    \includegraphics[width=\linewidth]{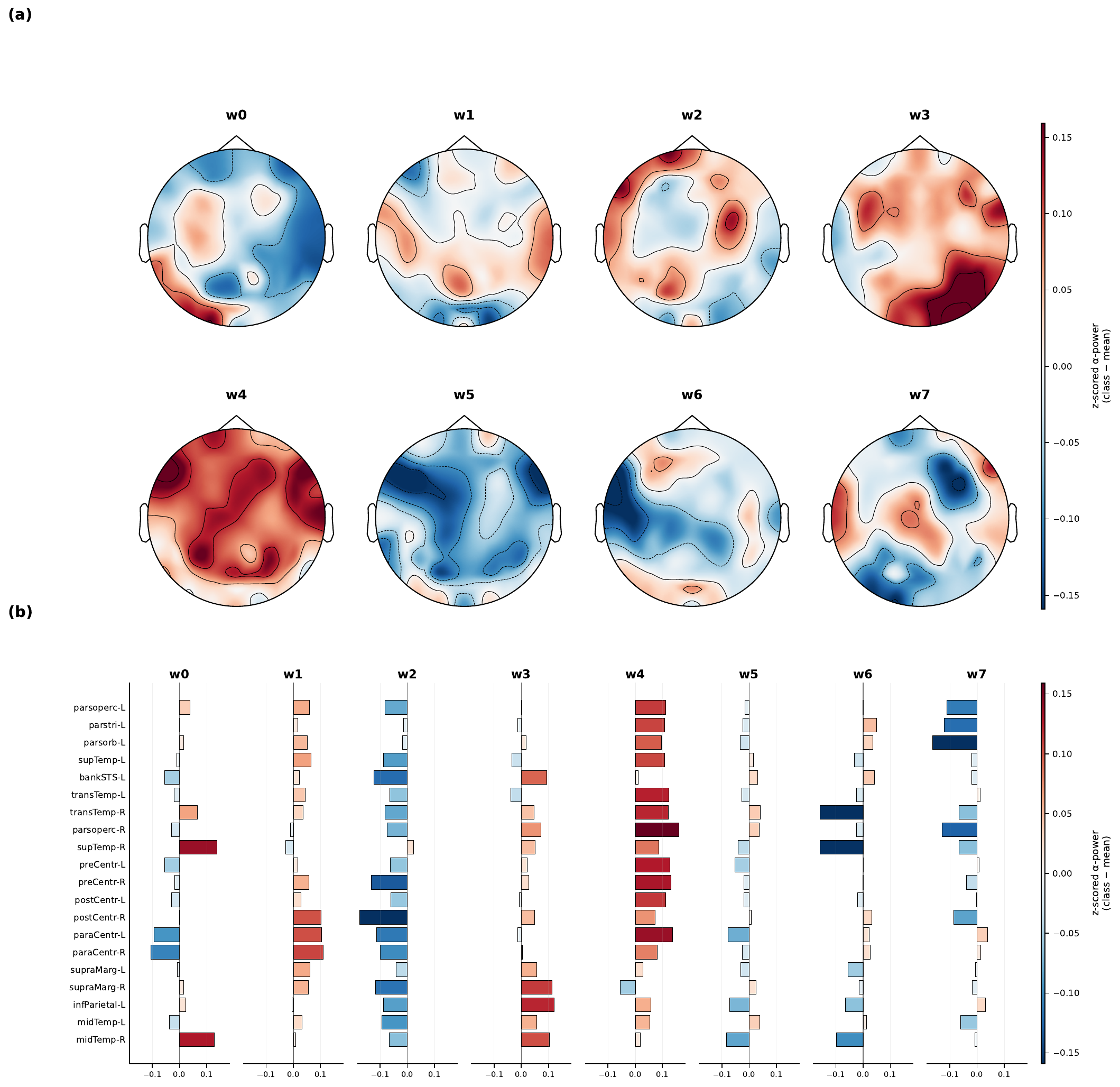}
    \caption{BISpeech (ds004196, $4$ subjects, $8$ semantic words: $4$ social $+$ $4$ numerical).}
    \label{fig:story_bispeech}
  \end{subfigure}
  \caption{Class-conditional alpha-band ($8$--$13$\,Hz) log-power signature on the two corpora with sufficient cohort to support a dedicated sensor + source panel: ToL (a) and BISpeech (b). In each subfigure, the top row shows per-class sensor-space scalp topomaps of the cross-subject grand-mean $z$-scored alpha power, centred on the across-class mean; the bottom row shows the matching source-space signature as a diverging horizontal-bar chart per class over the $20$ Desikan--Killiany language-network ROIs, where each bar extends left of the zero-line for negative deviations and right for positive deviations, on the same ``class $-$ mean'' colour scale. The class-differentiated fronto-temporal-parietal alpha pattern visible in (a) recurs in (b) despite the smaller cohort and the different (semantic) stimulus structure, with the social and numerical sub-vocabularies recruiting overlapping but distinguishable regions. The third corpus (SyncSpeech) is rendered together with the other two on the cortex in Figure~\ref{fig:brain3d_all}.}
  \label{fig:story_combined}
\end{figure}

\begin{figure}[H]
  \centering
  \begin{subfigure}[t]{0.55\linewidth}
    \centering
    \includegraphics[width=\linewidth]{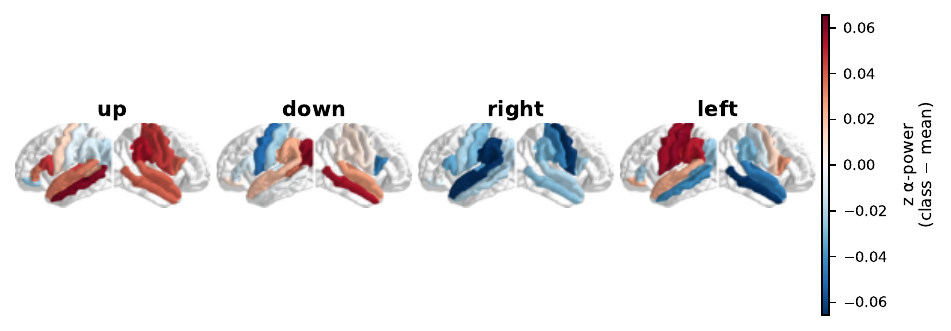}
    \caption{ToL (ds003626, $10$ subjects, $4$ directional classes: \textit{up}, \textit{down}, \textit{right}, \textit{left}).}
    \label{fig:brain3d_tol}
  \end{subfigure}\\[2pt]
  \begin{subfigure}[t]{\linewidth}
    \centering
    \includegraphics[width=\linewidth]{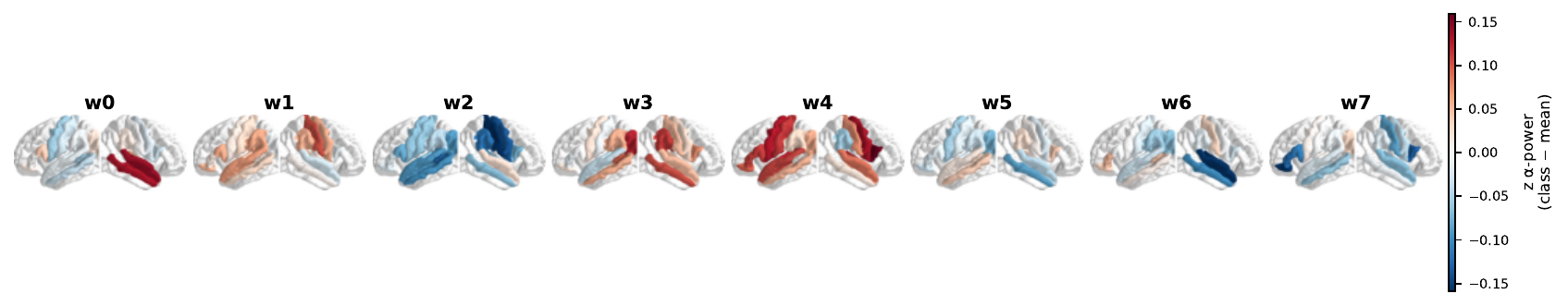}
    \caption{BISpeech (ds004196, $4$ subjects, $8$ semantic words: $4$ social $+$ $4$ numerical).}
    \label{fig:brain3d_bispeech}
  \end{subfigure}\\[2pt]
  \begin{subfigure}[t]{\linewidth}
    \centering
    \includegraphics[width=\linewidth]{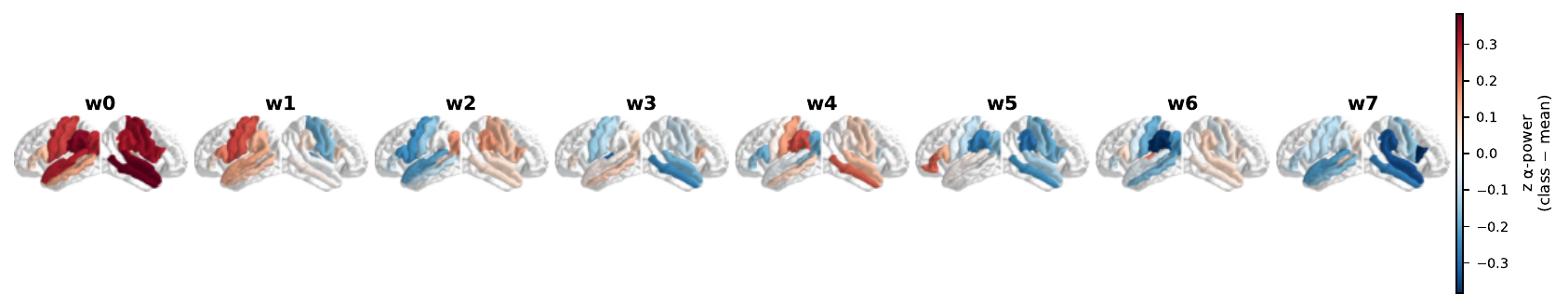}
    \caption{SyncSpeech ($3$ subjects, $8$ semantic words; same vocabulary as BISpeech, independent cohort, simultaneous EEG--fMRI acquisition).}
    \label{fig:brain3d_ds006033}
  \end{subfigure}
  \caption{Cortical projection of the class-conditional alpha-band signature onto the fsaverage pial surface for all three inner-speech corpora (left and right hemispheres, lateral views per class). The $20$ Desikan--Killiany language-network ROIs are coloured by their per-class deviation from the across-class mean, on the same scale used by the matching source-space diverging-bar charts for ToL and BISpeech in Figure~\ref{fig:story_combined} (panels~\ref{fig:story_tol} and~\ref{fig:story_bispeech}, bottom row of each); SyncSpeech is shown only on the cortex here. Reading down the figure makes the cross-dataset replication direct: in every corpus the class-conditional structure is carried by bilateral inferior frontal, superior temporal, and parietal regions of the cortical language network, despite differences in cohort size ($10$ / $4$ / $3$), task ($4$-class directional vs.\ $8$-word semantic), and acquisition hardware.}
  \label{fig:brain3d_all}
\end{figure}

The two-regime effect (\hyperref[subsec:innerspeech_results]{Inner-Speech Decoding Across Three Datasets}) and the cross-method ceiling (\hyperref[subsec:ceiling]{A Signal-Limited Accuracy Ceiling}) establish that sensor- and source-space decoding are complementary across architectures and that the ceiling is signal-limited rather than model-limited. To make the underlying signal directly visible, we compute for each dataset the class-conditional alpha-band ($8$--$13$\,Hz) log-power deviation from the grand mean, per subject and pooled across the cohort. The same quantity is rendered in two complementary forms below. The first (Figure~\ref{fig:story_combined}, panels~\ref{fig:story_tol} and~\ref{fig:story_bispeech} for ToL and BISpeech respectively) pairs a sensor-space scalp topomap per class (top of each panel) with a diverging horizontal-bar chart per class over the $20$ Desikan--Killiany language-network ROIs (bottom of each panel), where each bar extends left of the zero-line for negative deviations and right for positive deviations from the across-class mean. The third corpus (SyncSpeech) is not duplicated here in the same form because the source-space signature for all three datasets is rendered together on the cortex in Figure~\ref{fig:brain3d_all}, which carries the cross-dataset replication argument. The second (Figure~\ref{fig:brain3d_all}, panels (a)--(c)) projects the same source-space signature onto the fsaverage cortical surface, stacking the three datasets vertically so that the cross-dataset replication of the anatomical pattern is read down the page at a glance. Across all three corpora the alpha-band pattern is class-differentiated: each class drives a distinct combination of frontal, temporal, and parietal cortical regions, consistent with the cortical language network~\cite{PerroneBertolotti2014}. The recurrence of this structure across three cohorts of different size, hardware, and stimulus type is evidence that the discriminative signal is anatomically real and is not a per-dataset artefact.

\FloatBarrier

\subsection*{Learned Filter Spectrum}
\label{subsec:filter_spectrum}

\begin{figure}[H]
  \centering
  \includegraphics[width=\linewidth]{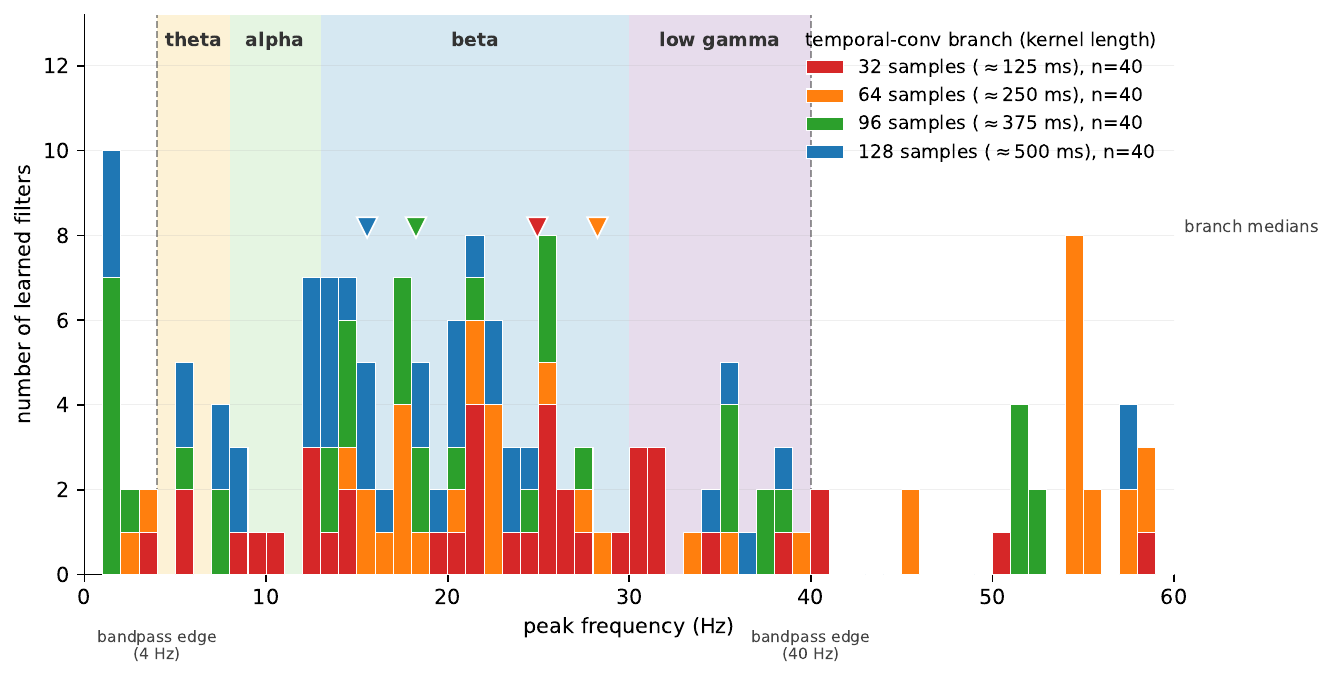}
  \caption{Peak-frequency distribution of VC-DATNet's learned temporal-conv filters. For each of the four kernel-length branches ($k \in \{32, 64, 96, 128\}$ samples at $256$\,Hz; $\approx 125$/$250$/$375$/$500$\,ms), the $8$ filters from the best-fold checkpoint of every trained target (ToL sub-01 sensor, plus sub-01/04/06/10 source; $40$ filters per branch, $160$ total) are zero-padded, FFT'd, and assigned their peak frequency over $1$--$60$\,Hz. Coloured bands shade the canonical EEG ranges (theta $4$--$8$, alpha $8$--$13$, beta $13$--$30$, low gamma $30$--$40$\,Hz); triangles mark the per-branch medians. The mass of the distribution lies in beta. The ${>}40$\,Hz tail is outside the $4$--$40$\,Hz training bandpass (dashed lines) and is not interpretable as a learned high-frequency feature.}
  \label{fig:filter_freq}
\end{figure}

The multi-scale temporal-convolution front end of VC-DATNet learns $32$ filters per stream, organised into four kernel-length branches ($\{32, 64, 96, 128\}$ samples at $256$\,Hz, $\approx \{125, 250, 375, 500\}$\,ms; \hyperref[subsubsec:multiscale_temporal]{Multi-scale Temporal Convolution}). Pooling the best-fold checkpoint from each of the five trained targets ($40$ filters per branch, $160$ total) and computing each filter's peak frequency over $1$--$60$\,Hz (Figure~\ref{fig:filter_freq}) places the mass of the distribution in the beta band ($13$--$30$\,Hz): the median peak per branch shifts from $24.9$\,Hz for the shortest kernel ($k\!=\!32$) to $15.6$\,Hz for the longest ($k\!=\!128$), so shorter kernels favour higher beta and longer kernels favour the alpha--low-beta boundary. The beta-band concentration is consistent with the role of sensorimotor and inner-speech beta rhythms in articulatory planning~\cite{PerroneBertolotti2014,nieto2022thinking,imaginedspeech_review2025}.

\FloatBarrier

\subsection*{Parameter Count}\phantomsection \label{subsec:comp_eff}

VC-DATNet is compact by design: in the canonical ToL configuration ($C\!=\!128$, $T\!=\!641$, $K\!=\!4$) it has $131.3$\,K trainable parameters (Table~\ref{tab:ceiling_full}), $\approx$$190\times$ fewer than the EEGPT foundation model ($25.0$\,M), $\approx$$2.6\times$ fewer than DeepConvNet ($345.9$\,K), and within an order of magnitude of the smallest baseline (EEGNet, $4.4$\,K). The exact count varies by a few thousand parameters with the input channel count and class count, both fixed by the dataset; most of the budget resides in the two dual-attention Transformer encoders, with the multi-scale temporal front-end and the linear classifier accounting for the remainder. We report the parameter count as a measure of architectural size only; FLOP, latency, and memory figures are not measured here and are deliberately not estimated.

\section*{Discussion}

This study used VC-DATNet, a compact dual-attention Transformer, as a single fixed instrument to decode inner speech from EEG across three independent public datasets (Thinking Out Loud, ds003626; the EEG modality of BISpeech, ds004196; and SyncSpeech, ds006033) under one preprocessing and evaluation protocol. Holding the architecture and protocol constant across datasets, and across the sensor- and source-space representations, was a deliberate methodological choice: it ensures that the differences reported below are properties of the data rather than of model selection.

\paragraph{Performance in context.}
On the primary ToL benchmark, under the cross-architecture bench's standard protocol (full $128$-channel montage, vanilla single-condition training; Table~\ref{tab:ceiling_full}), VC-DATNet reaches $37.58\%$ in sensor space, $35.11\%$ in source space, and $39.48\%$ as a per-subject sensor/source hybrid, all on $10$ subjects under within-subject $5$-fold cross-validation with seed $0$. The hybrid exceeds the strongest peer-reviewed baseline on this corpus ($36.1\%$ Gasparini et al.\ BiLSTM~\cite{Gasparini2022InnerSpeech}) by $3.4$ percentage points at approximately $43\%$ of its parameter budget ($131.3$\,K vs.\ $305.6$\,K), and is approximately $+7.9$ percentage points over the earlier EEGNet deep-learning baseline ($29.7\%$~\cite{vandenberg2021inner}). A sensitivity run on VC-DATNet's a priori $22$-channel inner-speech-relevant subset with cross-condition (Pron+Inner+Vis) multi-task pretrain followed by inner-only fine-tune reaches $34.42\%$ in sensor space and $36.95\%$ as the per-subject hybrid on the same $10$ subjects; the result is therefore robust to the choice of channel subset and to the cross-condition pretrain. On BISpeech the $8$-word accuracy ($20.60\%$) exceeds the strongest EEG-only baseline by $5.82$ percentage points (Wilson et al.\ random forest, $14.78\%$~\cite{wilson2023performance}); on SyncSpeech we report the first published decoding result ($25.53\%$ hybrid, chance $12.5\%$). All three accuracies lie within the leak-free envelope for inner-speech decoding (Table~\ref{tab:related_innerspeech}); reports far above this envelope are, on inspection, traceable to evaluation leakage rather than to genuine decoding gains.

\paragraph{Two complementary decoding regimes.}
The central finding is that sensor-space and cortical source-space decoding are complementary: each is the better representation for a distinct subset of subjects, and neither dominates (Tables~\ref{tab:tol_persubject},~\ref{tab:bispeech},~\ref{tab:ds006033}). The pattern appears independently on all three datasets, which were collected by different groups with different hardware and stimuli, which makes it a property of inner-speech EEG rather than a single-dataset artefact. A plausible mechanism is volume conduction: for subjects whose cortical language-network activity is focal, the sensor montage already resolves it and the source projection onto a fixed set of regions of interest discards usable information; for subjects whose scalp signal is weak or spatially smeared, the source projection concentrates the dilute focal activity and recovers it. Because one fixed architecture decoded both representations, the split cannot be a model artefact. The practical implication is that an inner-speech decoder should not commit to a single signal representation: a per-subject choice of representation (the ``hybrid'' here) is the genuine performance envelope, and is what a deployed system should select during calibration.

\paragraph{A signal-limited ceiling.}
With the architecture and protocol fixed, ten deep EEG decoders spanning recurrent, convolutional, TCN, CNN--Transformer hybrid, and pretrained-foundation families, evaluated on the same three datasets under the same protocol, converged to the same accuracy band on each dataset (Table~\ref{tab:ceiling_full}). All ten methods fell within a narrow $\sim$5--6\,pp band on every dataset, with the 25-million-parameter pretrained foundation model (linear-probe) sitting within $\sim$2\,pp of the cluster centre. The cross-method convergence is consistent with a signal-limited decoding ceiling on these public datasets under the present preprocessing pipeline, rather than one set by the capacity or inductive bias of the decoder. The finding is negative but substantive: it bounds what architectural work on these specific datasets can be expected to deliver and identifies data-side levers (larger cohorts, richer paradigms, better acquisition) as the productive route to higher inner-speech decoding accuracy.

\paragraph{The role of the architecture.}
The variance-pooling stage is motivated by the band-power signature of event-related (de)synchronisation, a second-order quantity that mean pooling is invariant to; Lemma~\ref{lem:logvar_psd} makes the link between log-variance pooling and band-limited log-power precise, and Proposition~\ref{prop:simplex} characterises the pre-softmax value-gating used in the attention block. These are design motivations, not performance claims. Table~\ref{tab:ceiling_full} places VC-DATNet within the same narrow accuracy band as the other nine decoders on every dataset; the value of the architecture in this study lies in serving as a consistent, reproducible, $131.3$\,K-parameter instrument that enabled the controlled three-dataset comparison at architectural-search cost an order of magnitude below the larger CNN--Transformer baselines, and whose two formal properties give the design choices an interpretable footing rather than an empirical-only one.

\subsection*{Limitations and Scope}
\label{subsec:limitations}

We state the limitations explicitly.

The cohorts are small: ToL provides 10 subjects, BISpeech 4, SyncSpeech 3. Cross-dataset replication of the two-regime effect is strong evidence relative to the inner-speech literature, but the absolute statistical power for any single dataset is limited, and the three-subject SyncSpeech result in particular should be read as a first indication rather than a definitive number.

The architecture is competitive, not superior. VC-DATNet matches the strongest peer-reviewed ToL baseline within noise and matches the other method families in Table~\ref{tab:ceiling_full}; it does not beat them. Its reported advantages are parameter efficiency ($\approx$131K parameters) and breadth of evaluation, not headline accuracy.

Several analyses that would strengthen the paper are not yet included, and are deliberately omitted rather than estimated: a component- and statistical-pooling ablation isolating the contribution of the variance stream; multi-seed runs with paired significance tests and confidence intervals; and a subject-independent (leave-one-subject-out) evaluation. These are planned as direct extensions (see below); the present results are within-subject five-fold cross-validation throughout.

The variance--power identity (Lemma~\ref{lem:logvar_psd}) assumes wide-sense stationarity of the band-pass-filtered signal within each pooling window. EEG is non-stationary at trial timescales, so the lemma applies to a piecewise-stationary approximation; the approximation is standard~\cite{stoica2005spectral,lotte2018review} but its bias has not been measured directly.

Following the established convention for this benchmark family, the accuracy of a fold is reported as its best test accuracy across the training run; this is used consistently for VC-DATNet and for every comparator in Table~\ref{tab:ceiling_full} to keep the cross-method comparison fair. The convention carries a known upward bias on small folds and is a source of variance that a leakage-free protocol (early stopping on a held-out validation split, or final-epoch evaluation) would remove. We retain the convention here so that the headline numbers are directly comparable with prior published baselines, and we identify a leakage-free re-evaluation as a useful follow-up.

Finally, absolute accuracy remains well below the threshold for an unsupervised inner-speech command interface. This reflects the difficulty of the paradigm and the signal-limited ceiling discussed above, not a deficiency specific to this architecture; practical use would require calibration, adaptive thresholding, or a reduced command set.

\section*{Conclusion}

Inner-speech BCIs have been studied almost exclusively in fragmented, single-corpus settings, with each new architecture reported on one dataset under one protocol. This paper establishes a controlled cross-dataset framework for the paradigm: one fixed architecture, one fixed protocol, three independent public corpora, two signal representations, and ten cross-method baselines under a single training recipe. Three findings emerge, each independent of any leaderboard claim. (i) Sensor-space and cortical source-space decoding are complementary: each is the better representation for a distinct subset of subjects, and the split replicates on all three datasets (Tables~\ref{tab:tol_persubject},~\ref{tab:bispeech},~\ref{tab:ds006033}) and across nine architecturally distinct decoders (Table~\ref{tab:tworegime_replication}), so it is a property of inner-speech EEG and not of any one model. (ii) Ten deep decoders spanning three orders of magnitude in parameter count, five architectural families, and a $25$\,M-parameter pretrained foundation model converge to the same narrow accuracy band on every dataset (Table~\ref{tab:ceiling_full}), consistent with a signal-limited rather than capacity-limited decoding ceiling on these corpora. (iii) On ToL, VC-DATNet's per-subject hybrid reaches $39.48\%$ (sensor $37.58\%$, source $35.11\%$; $10$ subjects, $5$-fold CV, seed $0$), $3.4$ percentage points above the strongest peer-reviewed baseline ($36.1\%$, Gasparini et al.~\cite{Gasparini2022InnerSpeech}) at less than half its parameter budget, and the first published decoding result on SyncSpeech ($25.53\%$ hybrid) is reported here. The paper claims an improvement over the prior ToL baseline only under the single-seed protocol shared with every other architecture in the bench (Table~\ref{tab:ceiling_full}); multi-seed significance is identified as a direct follow-up.

Three directions follow. First, the within-paradigm two-regime finding identifies per-subject representation selection (or a model that fuses sensor- and source-space streams) as the natural next architectural target, with the bench in Table~\ref{tab:ceiling_full} as the reference comparison. Second, because the ceiling on these public corpora is signal-limited, the highest-yield axis for further progress is data-side (larger inner-speech cohorts, richer paradigms, and acquisition protocols that improve trial-level signal-to-noise) rather than further architectural search on existing datasets. Third, the controlled-protocol scaffolding used here (one architecture, one preprocessing, one training recipe, three independent corpora, two signal representations, per-run logs) generalises directly to other low-signal-to-noise EEG paradigms where within-paradigm cross-corpus replication remains the open question. The complete preprocessing pipeline, source-space projection, model code, and per-run training logs are released to support these directions. The intended contribution of this paper is a reusable reference point for honest cross-dataset evaluation of inner-speech decoders, with the controlled protocol and the released artefacts together forming that reference.

%% Declaration of competing interest
\section*{Declaration of competing interest}
The authors declare that they have no known competing financial interests or personal relationships that could have appeared to influence the work reported in this paper.

%% Acknowledgements
\section*{Acknowledgements}
The authors thank the Computer Science Department at the American International University-Bangladesh (AIUB) for providing the necessary computational resources and research facilities. The study uses publicly available datasets with prior ethical approval as reported in the original publications \cite{nieto2022thinking, liwicki2023openneuro, simistira2025ds006033}. No new human data was collected.

%% Funding
\section*{Funding}
This research received no external funding.

%% Data availability
\section*{Data availability}
The three datasets analyzed in this study are publicly available from OpenNeuro.
Thinking Out Loud (ToL) dataset (ds003626): \url{https://openneuro.org/datasets/ds003626}.
Bimodal Inner Speech (BISpeech) dataset (ds004196): \url{https://openneuro.org/datasets/ds004196}.
SyncSpeech (Synchronous EEG--fMRI Inner Speech; ds006033): \url{https://openneuro.org/datasets/ds006033}.
The complete source code for the VC-DATNet model, the source-space projection, and the training and evaluation scripts is available at \url{https://github.com/wshuv-o/VC-DATNet}. The repository is open source under the MIT license and will remain accessible after publication.

%% Bibliography
\bibliography{cas-refs}

@article{lotte2018review,
title={A review of classification algorithms for EEG-based brain–computer interfaces: a 10 year update}, volume={15}, ISSN={1741-2552}, url={http://dx.doi.org/10.1088/1741-2552/aab2f2}, DOI={10.1088/1741-2552/aab2f2}, number={3}, journal={Journal of Neural Engineering}, publisher={IOP Publishing}, author={Lotte, F and Bougrain, L and Cichocki, A and Clerc, M and Congedo, M and Rakotomamonjy, A and Yger, F}, year={2018}, month=apr, pages={031005} }

@book{riyad-2020,
	author = {Riyad, Mouad and Khalil, Mohammed and Adib, Abdellah},
	booktitle = {Lecture notes in computer science},
	month = {1},
	pages = {103--111},
	title = {{Incep-EEGNET: a ConVNet for motor imagery decoding}},
	year = {2020},
	doi = {10.1007/978-3-030-51935-3_11},
	url = {https://doi.org/10.1007/978-3-030-51935-3_11},
}

@article{schirrmeister2017deep,
title={Deep learning with convolutional neural networks for EEG decoding and visualization}, volume={38}, ISSN={1097-0193}, url={http://dx.doi.org/10.1002/hbm.23730}, DOI={10.1002/hbm.23730}, number={11}, journal={Human Brain Mapping}, publisher={Wiley}, author={Schirrmeister, Robin Tibor and Springenberg, Jost Tobias and Fiederer, Lukas Dominique Josef and Glasstetter, Martin and Eggensperger, Katharina and Tangermann, Michael and Hutter, Frank and Burgard, Wolfram and Ball, Tonio}, year={2017}, month=aug, pages={5391–5420} }

@inproceedings{Gasparini2022InnerSpeech,
  author    = {Francesca Gasparini and Elisa Cazzaniga and Aurora Saibene},
  title     = {Inner speech recognition through electroencephalographic signals},
  booktitle = {Proceedings of the 1st Workshop on Artificial Intelligence for Human Machine Interaction (AIxHMI 2022)},
  pages     = {48--61},
  year      = {2022},
  editor    = {Aurora Saibene and Silvia Corchs and Jordi Solé-Casals},
  volume    = {3368},
  series    = {CEUR Workshop Proceedings},
  publisher = {CEUR-WS.org},
  address   = {Udine, Italy},
  url       = {https://boa.unimib.it/handle/10281/394066}
}

@article{zhang2024eegmamba,
  title={EEGMamba: Bidirectional State Space Model with Mixture of Experts for EEG Classification},
  author={Zhang, Yifan and Wang, Jun and Li, Xiaoyu and Liu, Yujie},
  journal={arXiv preprint arXiv:2407.20254},
  year={2024}
}

@inproceedings{ingolfsson2020eeg,
  title={EEG-TCNet: An Accurate Temporal Convolutional Network for Embedded Motor-Imagery Brain–Machine Interfaces},
  author={Ingolfsson, Thorsteinn M. and Hersche, Michael and Wang, Xin and Cavigelli, Lukas},
  booktitle={Proceedings of the 2020 IEEE International Conference on Systems, Man, and Cybernetics (SMC)},
  pages={2958--2965},
  year={2020},
  organization={IEEE}
}

@article{ma2024attention,
title={Attention-based convolutional neural network with multi-modal temporal information fusion for motor imagery EEG decoding}, volume={175}, ISSN={0010-4825}, url={http://dx.doi.org/10.1016/j.compbiomed.2024.108504}, DOI={10.1016/j.compbiomed.2024.108504}, journal={Computers in Biology and Medicine}, publisher={Elsevier BV}, author={Ma, Xinzhi and Chen, Weihai and Pei, Zhongcai and Zhang, Yue and Chen, Jianer}, year={2024}, month=jun, pages={108504} }

@article{lawhern2018eegnet,
title={EEGNet: a compact convolutional neural network for EEG-based brain–computer interfaces}, volume={15}, ISSN={1741-2552}, url={http://dx.doi.org/10.1088/1741-2552/aace8c}, DOI={10.1088/1741-2552/aace8c}, number={5}, journal={Journal of Neural Engineering}, publisher={IOP Publishing}, author={Lawhern, Vernon J and Solon, Amelia J and Waytowich, Nicholas R and Gordon, Stephen M and Hung, Chou P and Lance, Brent J}, year={2018}, month=jul, pages={056013} }

@article{craik2019deep,
title={Deep learning for electroencephalogram (EEG) classification tasks: a review}, volume={16}, ISSN={1741-2552}, url={http://dx.doi.org/10.1088/1741-2552/ab0ab5}, DOI={10.1088/1741-2552/ab0ab5}, number={3}, journal={Journal of Neural Engineering}, publisher={IOP Publishing}, author={Craik, Alexander and He, Yongtian and Contreras-Vidal, Jose L}, year={2019}, month=apr, pages={031001} }

@article{roy2019chronnectome,
title={Chronnectome fingerprinting: Identifying individuals and predicting higher cognitive functions using dynamic brain connectivity patterns}, volume={39}, ISSN={1097-0193}, url={http://dx.doi.org/10.1002/hbm.23890}, DOI={10.1002/hbm.23890}, number={2}, journal={Human Brain Mapping}, publisher={Wiley}, author={Liu, Jin and Liao, Xuhong and Xia, Mingrui and He, Yong}, year={2017}, month=nov, pages={902–915} }

@article{vafaei2025transformers,
AUTHOR = {Vafaei, Elnaz and Hosseini, Mohammad},
TITLE = {Transformers in EEG Analysis: A Review of Architectures and Applications in Motor Imagery, Seizure, and Emotion Classification},
JOURNAL = {Sensors},
VOLUME = {25},
YEAR = {2025},
NUMBER = {5},
ARTICLE-NUMBER = {1293},
URL = {https://www.mdpi.com/1424-8220/25/5/1293},
PubMedID = {40096020},
ISSN = {1424-8220},
DOI = {10.3390/s25051293}
}

@article{Gramfort2013,
title={MEG and EEG data analysis with MNE-Python}, volume={7}, ISSN={1662-453X}, url={http://dx.doi.org/10.3389/fnins.2013.00267}, DOI={10.3389/fnins.2013.00267}, journal={Frontiers in Neuroscience}, publisher={Frontiers Media SA}, author={Gramfort, Alexandre}, year={2013} }

@article{ILLE2024149,
title = {Ongoing EEG artifact correction using blind source separation},
journal = {Clinical Neurophysiology},
volume = {158},
pages = {149-158},
year = {2024},
issn = {1388-2457},
doi = {https://doi.org/10.1016/j.clinph.2023.12.133},
url = {https://www.sciencedirect.com/science/article/pii/S1388245723009380},
author = {Nicole Ille and Yoshiaki Nakao and Shumpei Yano and Toshiyuki Taura and Arndt Ebert and Harald Bornfleth and Suguru Asagi and Kanoko Kozawa and Izumi Itabashi and Takafumi Sato and Rie Sakuraba and Rie Tsuda and Yosuke Kakisaka and Kazutaka Jin and Nobukazu Nakasato}
}

@article{Delorme2004,
 title={EEGLAB: an open source toolbox for analysis of single-trial EEG dynamics including independent component analysis}, volume={134}, ISSN={0165-0270}, url={http://dx.doi.org/10.1016/j.jneumeth.2003.10.009}, DOI={10.1016/j.jneumeth.2003.10.009}, number={1}, journal={Journal of Neuroscience Methods}, publisher={Elsevier BV}, author={Delorme, Arnaud and Makeig, Scott}, year={2004}, month=mar, pages={9–21} }

@article{Widmann2015,
title={Digital filter design for electrophysiological data – a practical approach}, volume={250}, ISSN={0165-0270}, url={http://dx.doi.org/10.1016/j.jneumeth.2014.08.002}, DOI={10.1016/j.jneumeth.2014.08.002}, journal={Journal of Neuroscience Methods}, publisher={Elsevier BV}, author={Widmann, Andreas and Schröger, Erich and Maess, Burkhard}, year={2015}, month=jul, pages={34–46} }

@article{PerroneBertolotti2014,
title={Inner speech: Development, cognitive functions, phenomenology, and neurobiology.}, volume={141}, ISSN={0033-2909}, url={http://dx.doi.org/10.1037/bul0000021}, DOI={10.1037/bul0000021}, number={5}, journal={Psychological Bulletin}, publisher={American Psychological Association (APA)}, author={Alderson-Day, Ben and Fernyhough, Charles}, year={2015}, month=sep, pages={931–965} }

@article{roy2019deep,
title={Deep learning-based electroencephalography analysis: a systematic review}, volume={16}, ISSN={1741-2552}, url={http://dx.doi.org/10.1088/1741-2552/ab260c}, DOI={10.1088/1741-2552/ab260c}, number={5}, journal={Journal of Neural Engineering}, publisher={IOP Publishing}, author={Roy, Yannick and Banville, Hubert and Albuquerque, Isabela and Gramfort, Alexandre and Falk, Tiago H and Faubert, Jocelyn}, year={2019}, month=aug, pages={051001} }

@article{kostas2021bendr,
  author    = {Kostas, D. and Rudzicz, F.},
  title     = {BENDR: using transformers and a contrastive self-supervised learning task to learn from massive amounts of EEG data},
  journal   = {Frontiers in Human Neuroscience},
  volume    = {15},
  pages     = {653659},
  year      = {2021}
}

@article{pan-2024,
	author = {Pan, Yue and Liu, Qile and Liu, Qing and Zhang, Li and Huang, Gan and Chen, Xin and Li, Fali and Xu, Peng and Liang, Zhen},
	journal = {arXiv (Cornell University)},
	month = {7},
	title = {{DUA: Dual Attentive Transformer in Long-Term Continuous EEG Emotion Analysis}},
	year = {2024},
	doi = {10.48550/arxiv.2407.20519},
	url = {http://arxiv.org/abs/2407.20519},
}

@article{ding-2024,
	author = {Ding, Yi and Li, Yong and Sun, Hao and Liu, Rui and Tong, Chengxuan and Liu, Chenyu and Zhou, Xinliang and Guan, Cuntai},
	journal = {IEEE Journal of Biomedical and Health Informatics},
	month = {11},
	number = {3},
	pages = {1909--1918},
	title = {{EEG-Deformer: a dense convolutional transformer for Brain-Computer interfaces}},
	volume = {29},
	year = {2024},
	doi = {10.1109/jbhi.2024.3504604},
	url = {https://doi.org/10.1109/jbhi.2024.3504604},
}

@article{zhao2024ctnet,
  title={CTNet: a convolutional transformer network for EEG-based motor imagery classification},
  author={Zhao, Wei and Jiang, Xiaolu and Zhang, Baocan and Xiao, Shixiao and Weng, Sujun and others},
  journal={Scientific Reports},
  volume={14},
  pages={20237},
  year={2024},
  doi={10.1038/s41598-024-71118-7}
}

@article{liu2025dmsacnn,
  title={DMSACNN: Deep Multiscale Attentional Convolutional Neural Network for EEG-Based Motor Decoding},
  author={Liu, Ke and Xing, Xin and Yang, Tao and Yu, Zhuliang and Xiao, Bin and Wang, Guoyin and Wu, Wei},
  journal={IEEE Journal of Biomedical and Health Informatics},
  year={2025},
  doi={10.1109/JBHI.2025.3546288}
}

@article{du2022dual,
  title={Dual attentive fusion for EEG-based brain–computer interfaces},
  author={Du, Yuanhua and Huang, Jian and Huang, Xiuyu and Shi, Kaibo and Zhou, Nan},
  journal={Frontiers in Neuroscience},
  volume={16},
  pages={1044631},
  year={2022},
  doi={10.3389/fnins.2022.1044631}
}

@article{tang2023motor,
  title={Motor Imagery EEG Decoding Based on Multi-Scale Hybrid Networks and Feature Enhancement},
  author={Tang, Xianlun and Yang, Caiquan and Sun, Xia and Zou, Mi and Wang, Huiming},
  journal={IEEE Transactions on Neural Systems and Rehabilitation Engineering},
  volume={31},
  pages={1208--1218},
  year={2023},
  doi={10.1109/TNSRE.2023.3242280}
}

@article{li2024dual,
  title={Dual-TSST: A Dual-Branch Temporal-Spectral-Spatial Transformer Model for EEG Decoding},
  author={Li, Hongqi and Zhang, Haodong and Chen, Yitong},
  journal={arXiv preprint arXiv:2409.03251},
  year={2024},
doi={https://doi.org/10.48550/arXiv.2409.03251}
}

@article{zhang2023local,
  title={Local and global convolutional transformer-based motor imagery EEG classification},
  author={Zhang, Jiayang and Li, Kang and Yang, Banghua and Han, Xiaofei},
  journal={Frontiers in Neuroscience},
  volume={17},
  pages={1219988},
  year={2023},
  doi={10.3389/fnins.2023.1219988}
}

@article{Kappenman2021,
title={ERP CORE: An open resource for human event-related potential research}, volume={225}, ISSN={1053-8119}, url={http://dx.doi.org/10.1016/j.neuroimage.2020.117465}, DOI={10.1016/j.neuroimage.2020.117465}, journal={NeuroImage}, publisher={Elsevier BV}, author={Kappenman, Emily S. and Farrens, Jaclyn L. and Zhang, Wendy and Stewart, Andrew X. and Luck, Steven J.}, year={2021}, month=jan, pages={117465} }

@article{xu-2024,
	author = {Xu, Wenyi and Monachino, Alexa D and McCormick, Sarah A and Margolis, Emma T and Sobrino, Ana and Bosco, Cara and Franke, Cassandra J and Davel, Lauren and Zieff, Michal R and Donald, Kirsten A and Gabard-Durnam, Laurel J and Morales, Santiago},
	journal = {Developmental Cognitive Neuroscience},
	month = {9},
	pages = {101458},
	title = {{Advancing the reporting of pediatric EEG data: Tools for estimating reliability, effect size, and data quality metrics}},
	volume = {70},
	year = {2024},
	doi = {10.1016/j.dcn.2024.101458},
	url = {https://doi.org/10.1016/j.dcn.2024.101458},
}

@article{nieto2022thinking,
title={Thinking out loud, an open-access EEG-based BCI dataset for inner speech recognition}, volume={9}, ISSN={2052-4463}, url={http://dx.doi.org/10.1038/s41597-022-01147-2}, DOI={10.1038/s41597-022-01147-2}, number={1}, journal={Scientific Data}, publisher={Springer Science and Business Media LLC}, author={Nieto, Nicolás and Peterson, Victoria and Rufiner, Hugo Leonardo and Kamienkowski, Juan Esteban and Spies, Ruben}, year={2022}, month=feb }

@article{dong-2021,
	author = {Dong, Li and Zhao, Lingling and Zhang, Yufan and Yu, Xue and Li, Fali and Li, Jianfu and Lai, Yongxiu and Liu, Tiejun and Yao, Dezhong},
	journal = {Brain Topography},
	month = {5},
	number = {4},
	pages = {403--414},
	title = {{Reference Electrode Standardization Interpolation Technique (RESIT): a novel interpolation method for scalp EEG}},
	volume = {34},
	year = {2021},
	doi = {10.1007/s10548-021-00844-2},
	url = {https://doi.org/10.1007/s10548-021-00844-2},
}

@article{mane2021fbcnet,
  title={FBCNet: A Multi-view Convolutional Neural Network for Brain-Computer Interface},
  author={Mane, Ravikiran and Chew, Effie and Chua, Karen and Ang, Kai Keng and Robinson, Neethu and Vinod, A.P. and Lee, Seong-Whan and Guan, Cuntai},
  journal={arXiv preprint arXiv:2104.01233},
  year={2021},
  doi={10.48550/arXiv.2104.01233}
}

@article{wu2025ameegnet,
  title={AMEEGNet: attention-based multiscale EEGNet for effective motor imagery EEG decoding},
  author={Wu, Xuejian and Chu, Yaqi and Li, Qing and Luo, Yang and Zhao, Yiwen and Zhao, Xingang},
  journal={Frontiers in Neurorobotics},
  volume={19},
  pages={1540033},
  year={2025},
  doi={10.3389/fnbot.2025.1540033},
  url={https://www.frontiersin.org/journals/neurorobotics/articles/10.3389/fnbot.2025.1540033/full}
}

@article{liao2025ciacnet,
  title={A composite improved attention convolutional network for motor imagery EEG classification},
  author={Liao, Wenzhe and Miao, Zipeng and Liang, Shuaibo and Zhang, Linyan and Li, Chen},
  journal={Frontiers in Neuroscience},
  volume={19},
  pages={1543508},
  year={2025},
  doi={10.3389/fnins.2025.1543508},
  url={https://www.frontiersin.org/journals/neuroscience/articles/10.3389/fnins.2025.1543508/full}
}

@article{wang2025amanet,
  title={AMANet: a data-augmented multi-scale temporal attention convolutional network for motor imagery classification},
  author={Wang, Shu and Wang, Raofen and Chang, Liang and Wu, Jianzhen and Hu, Lingyan},
  journal={Frontiers in Neurorobotics},
  volume={19},
  pages={1704111},
  year={2025},
  doi={10.3389/fnbot.2025.1704111},
  url={https://www.frontiersin.org/journals/neurorobotics/articles/10.3389/fnbot.2025.1704111/full}
}

@article{xie2024bfatcnet,
  title={Bidirectional feature pyramid attention-based temporal convolutional network model for motor imagery electroencephalogram classification},
  author={Xie, Xinghe and Chen, Liyan and Qin, Shujia and Zha, Fusheng and Fan, Xinggang},
  journal={Frontiers in Neurorobotics},
  volume={18},
  pages={1343249},
  year={2024},
  doi={10.3389/fnbot.2024.1343249},
  url={https://www.frontiersin.org/journals/neurorobotics/articles/10.3389/fnbot.2024.1343249/full}
}

@article{keutayeva2024eegcct,
  title={Compact convolutional transformer for subject-independent motor imagery EEG-based BCIs},
  author={Keutayeva, Aigerim and Fakhrutdinov, Nail and Abibullaev, Berdakh},
  journal={Scientific Reports},
  volume={14},
  pages={25775},
  year={2024},
  doi={10.1038/s41598-024-73755-4},
  url={https://www.nature.com/articles/s41598-024-73755-4}
}

@article{liao2025eegencoder,
  title={Advancing BCI with a transformer-based model for motor imagery classification},
  author={Liao, Jiahao and Liu, Xinyu and Wang, Fei},
  journal={Scientific Reports},
  volume={15},
  year={2025},
  doi={10.1038/s41598-025-06364-4},
  url={https://www.nature.com/articles/s41598-025-06364-4}
}

@article{satransnet2025,
  title={{SATrans-Net}: Sparse Attention Transformer for {EEG}-based motor imagery decoding},
  author={Miao, Tianhua and Sha, Liansen and Huang, Kun and Li, Yongbin and Liu, Bin},
  journal={Scientific Reports},
  volume={15},
  year={2025},
  doi={10.1038/s41598-025-30806-8},
  url={https://www.nature.com/articles/s41598-025-30806-8}
}

@article{sstaf2025,
  title={{SSTAF}: Spatial-Spectral-Temporal Attention Fusion Transformer for Motor Imagery Classification},
  author={Muna, Ummay Maria and Shawon, Md Mehedi Hasan and Jobayer, Md and Akter, Sumaiya and Sabuj, Saifur Rahman},
  journal={IEEE Transactions on Neural Systems and Rehabilitation Engineering},
  year={2025},
  doi={10.1109/TNSRE.2025.3560428},
  url={https://arxiv.org/abs/2504.13220}
}

@article{altaheri2025tcformer,
  title={Temporal convolutional transformer for EEG based motor imagery decoding},
  author={Altaheri, Hamdi and Karray, Fakhri and Karimi, Amir-Hossein},
  journal={Scientific Reports},
  volume={15},
  year={2025},
  doi={10.1038/s41598-025-16219-7},
  url={https://www.nature.com/articles/s41598-025-16219-7}
}

@article{zhao2025mscformer,
  title={Multi-scale convolutional transformer network for motor imagery brain-computer interface},
  author={Zhao, Wei and Zhang, Baocan and Zhou, Haifeng and Wei, Dezhi and Huang, Chenxi and Lan, Quan},
  journal={Scientific Reports},
  volume={15},
  pages={12935},
  year={2025},
  doi={10.1038/s41598-025-96611-5},
  url={https://www.nature.com/articles/s41598-025-96611-5}
}

@article{cltnet2025,
  title={{CLTNet}: A hybrid deep learning model for motor imagery classification},
  author={Gu, He and Chen, Ting-Wei and Ma, Xiao and Zhang, Mengyuan and Sun, Yan and Zhao, Jian},
  journal={Brain Sciences},
  volume={15},
  number={2},
  pages={124},
  year={2025},
  doi={10.3390/brainsci15020124},
  url={https://doi.org/10.3390/brainsci15020124}
}

@article{han2024edpnet,
  title={A Spatial-Spectral and Temporal Dual Prototype Network for Motor Imagery Brain-Computer Interface},
  author={Han, Can and Liu, Chen and Wang, Yaqi and Cai, Crystal and Wang, Jun and Qian, Dahong},
  journal={arXiv preprint arXiv:2407.03177},
  year={2024},
  doi={10.48550/arXiv.2407.03177},
  url={https://arxiv.org/abs/2407.03177}
}

@article{chen2024damgcn,
  title={EEG-based emotion recognition using graph convolutional neural network with dual attention mechanism},
  author={Chen, Wei and Liao, Yuan and Dai, Rui and Dong, Yuanlin and Huang, Liya},
  journal={Frontiers in Computational Neuroscience},
  volume={18},
  pages={1416494},
  year={2024},
  doi={10.3389/fncom.2024.1416494},
  url={https://www.frontiersin.org/journals/computational-neuroscience/articles/10.3389/fncom.2024.1416494/full}
}

@inproceedings{yang2024stmambanet,
  title={Spatial-Temporal Mamba Network for EEG-based Motor Imagery Classification},
  author={Yang, Xiaoxiao and Jia, Ziyu},
  booktitle={Advanced Data Mining and Applications (ADMA 2024)},
  year={2024},
  doi={10.48550/arXiv.2409.09627},
  url={https://arxiv.org/abs/2409.09627}
}

@article{milyani2025innerspeech,
  title={Deep learning for inner speech recognition: a pilot comparative study of EEGNet and a spectro-temporal Transformer on bimodal EEG-fMRI data},
  author={Milyani, Ahmad H and Attar, Eyad Talal},
  journal={Frontiers in Human Neuroscience},
  volume={19},
  pages={1668935},
  year={2025},
  doi={10.3389/fnhum.2025.1668935},
  url={https://www.frontiersin.org/journals/human-neuroscience/articles/10.3389/fnhum.2025.1668935/full}
}

@article{innerspeech_future2025,
  title={Neural signals, machine learning, and the future of inner speech recognition},
  author={Chowdhury, Adiba Tabassum and Hassanein, Ahmed and Al~Shibli, Aous N. and Khanafer, Youssuf and AbuHaweeleh, Mohannad and Pedersen, Shona and Chowdhury, Muhammad E. H.},
  journal={Frontiers in Human Neuroscience},
  volume={19},
  pages={1637174},
  year={2025},
  doi={10.3389/fnhum.2025.1637174},
  url={https://www.frontiersin.org/journals/human-neuroscience/articles/10.3389/fnhum.2025.1637174/full}
}

@article{imaginedspeech_review2025,
  title={Systematic review: progress in {EEG}-based speech imagery brain--computer interface decoding and encoding research},
  author={Su, Ke and Tian, Liang},
  journal={PeerJ Computer Science},
  volume={11},
  pages={e2938},
  year={2025},
  doi={10.7717/peerj-cs.2938},
  url={https://peerj.com/articles/cs-2938/}
}

@inproceedings{woo2018cbam,
  title={CBAM: Convolutional Block Attention Module},
  author={Woo, Sanghyun and Park, Jongchan and Lee, Joon-Young and Kweon, In So},
  booktitle={Proceedings of the European Conference on Computer Vision (ECCV)},
  pages={3--19},
  year={2018},
  doi={10.1007/978-3-030-01234-2_1}
}

@book{stoica2005spectral,
  title={Spectral Analysis of Signals},
  author={Stoica, Petre and Moses, Randolph L.},
  year={2005},
  publisher={Pearson Prentice Hall},
  address={Upper Saddle River, NJ}
}

@article{simistira2023bimodal,
  title={Bimodal electroencephalography-functional magnetic resonance imaging dataset for inner-speech recognition},
  author={Simistira Liwicki, Foteini and Gupta, Vibha and Saini, Rajkumar and De, Kanjar and Abid, Nosheen and Rakesh, Sumit and Wellington, Scott and Wilson, Holly and Liwicki, Marcus and Eriksson, Johan},
  journal={Scientific Data},
  volume={10},
  pages={378},
  year={2023},
  publisher={Nature Publishing Group},
  doi={10.1038/s41597-023-02286-w},
  url={https://www.nature.com/articles/s41597-023-02286-w}
}

@misc{liwicki2023openneuro,
  title={Bimodal dataset on Inner speech},
  author={Liwicki, Foteini and Gupta, Vibha and Saini, Rajkumar and De, Kanjar and Abid, Nosheen and Rakesh, Sumit and Wellington, Scott and Wilson, Holly and Liwicki, Marcus and Eriksson, Johan},
  year={2023},
  howpublished={OpenNeuro},
  doi={10.18112/openneuro.ds004196.v2.0.2},
  note={Dataset, accession ds004196}
}

@article{wilson2022bimodal,
  title={Bimodal pilot study on inner speech decoding reveals the potential of combining {EEG} and {fMRI}},
  author={Wilson, Holly and Wellington, Scott and Simistira Liwicki, Foteini and Gupta, Vibha and Saini, Rajkumar and De, Kanjar and Abid, Nosheen and Rakesh, Sumit and Eriksson, Johan and Watts, Oliver and Chen, Xiang and Golbabaee, Mohammad and Proulx, Michael J. and Liwicki, Marcus and O'Neill, Eamonn and Metcalfe, Benjamin},
  journal={bioRxiv},
  year={2022},
  doi={10.1101/2022.05.24.492109},
  url={https://www.biorxiv.org/content/10.1101/2022.05.24.492109v1}
}

@inproceedings{wilson2023performance,
  title={Performance of data-driven inner speech decoding with same-task {EEG--fMRI} data fusion and bimodal models},
  author={Wilson, Holly and Wellington, Scott and Simistira Liwicki, Foteini and Gupta, Vibha and Saini, Rajkumar and De, Kanjar and Abid, Nosheen and Rakesh, Sumit and Eriksson, Johan and Watts, Oliver and Chen, Xiang and Golbabaee, Mohammad and Proulx, Michael J. and Liwicki, Marcus and O'Neill, Eamonn and Metcalfe, Benjamin},
  booktitle={Proceedings of INTERSPEECH 2023},
  year={2023},
  eprint={2306.10854},
  archivePrefix={arXiv},
  url={https://arxiv.org/abs/2306.10854}
}

@inproceedings{wellington2024improving,
  title={Improving inner speech decoding by hybridisation of bimodal {EEG} and {fMRI} data},
  author={Wellington, Scott and Wilson, Holly and Simistira Liwicki, Foteini and Gupta, Vibha and Saini, Rajkumar and De, Kanjar and Abid, Nosheen and Rakesh, Sumit and Eriksson, Johan and Watts, Oliver and Chen, Xiang and Coyle, Damien and Golbabaee, Mohammad and Proulx, Michael J. and Liwicki, Marcus and O'Neill, Eamonn and Metcalfe, Benjamin},
  booktitle={2024 46th Annual International Conference of the IEEE Engineering in Medicine and Biology Society (EMBC)},
  pages={1--5},
  year={2024},
  organization={IEEE},
  doi={10.1109/EMBC53108.2024.10781692}
}

@article{qin2024crossperception,
  title={Exploring inner speech recognition via cross-perception approach in {EEG} and {fMRI}},
  author={Qin, Jianwei and Zong, Lin and Liu, Feng},
  journal={Applied Sciences},
  volume={14},
  number={17},
  pages={7720},
  year={2024},
  publisher={MDPI},
  doi={10.3390/app14177720},
  url={https://www.mdpi.com/2076-3417/14/17/7720}
}

@misc{abdalla2025multibranch,
  title={Inner speech {EEG} classification via a feature-fused multi-branch deep network: findings from an ablation analysis},
  author={Abdalla, Hesham E. M. and Basri, Hamza and Aris, Ishak and Khan, Atif H. and Natarajan, Sundararajan and Adha, Mohamad S. and Saga, Ahmad and Abbasa, Wael F. and Al-Haddad, Syed A. R.},
  year={2025},
  howpublished={SSRN preprint},
  note={Abstract id 5822199},
  url={https://papers.ssrn.com/sol3/papers.cfm?abstract_id=5822199}
}

@article{abdalla2025plvcsp,
  title={{EEG}-Based Inner Speech Decoding Using Phase-Locking and Spatial Features with a Dual-Branch Deep Learning Model},
  author={Abdalla, Hussna Elnoor and Basri, Hamidon and Aris, Ishak and Al-Haddad, Syed Abdul Rahman},
  journal={Open Research Africa},
  volume={8},
  pages={26},
  year={2025},
  doi={10.12688/openresafrica.16183.1},
  url={https://openresearchafrica.org/articles/8-26}
}

@article{aguilar2025spatial,
  title={Spatial Dynamics and Functional Connectivity in {EEG}: Insights from Lexical Processing},
  author={Aguilar, Omar Mejia Hook and Grullon, Ramon},
  journal={bioRxiv},
  year={2025},
  doi={10.1101/2025.04.29.651245},
  url={https://www.biorxiv.org/content/10.1101/2025.04.29.651245v1},
  note={Within-subject classification reported only for sub-02; n=1 result}
}

@article{agarwal2025tetra,
  title={Motor imagery-based neural networks for assisting tetraplegic patients},
  author={Agarwal, Prabhakar and Kumar, Sandeep and Singh, Rishav},
  journal={Medical \& Biological Engineering \& Computing},
  volume={63},
  number={12},
  pages={3793--3807},
  year={2025},
  doi={10.1007/s11517-025-03433-8}
}

@article{simistira2025ds006033,
  title={Simultaneous electroencephalography (EEG) and functional magnetic resonance imaging (fMRI) data during an inner speech task},
  author={Simistira Liwicki, Foteini and Saini, Rajkumar and Das Chakladar, Debashis and Rakesh, Sumit and Gupta, Vibha and Liwicki, Marcus and Eriksson, Johan},
  journal={Data in Brief},
  volume={63},
  pages={112258},
  year={2025},
  publisher={Elsevier},
  doi={10.1016/j.dib.2025.112258}
}

@inproceedings{vandenberg2021inner,
  title={Inner speech classification using {EEG} signals: a deep learning approach},
  author={van den Berg, Bram and van Donkelaar, Sander and Alimardani, Maryam},
  booktitle={2021 IEEE 2nd International Conference on Human-Machine Systems (ICHMS)},
  pages={1--4},
  year={2021},
  organization={IEEE},
  doi={10.1109/ICHMS53169.2021.9582457}
}

@article{dasalla2009single,
  title={Single-trial classification of vowel speech imagery using common spatial patterns},
  author={DaSalla, Charles S and Kambara, Hiroyuki and Sato, Makoto and Koike, Yasuharu},
  journal={Neural Networks},
  volume={22},
  number={9},
  pages={1334--1339},
  year={2009},
  publisher={Elsevier},
  doi={10.1016/j.neunet.2009.05.008}
}

@inproceedings{zhao2015classifying,
  title={Classifying phonological categories in imagined and articulated speech},
  author={Zhao, Shunan and Rudzicz, Frank},
  booktitle={2015 IEEE International Conference on Acoustics, Speech and Signal Processing (ICASSP)},
  pages={992--996},
  year={2015},
  organization={IEEE},
  doi={10.1109/ICASSP.2015.7178118}
}

@inproceedings{coretto2017open,
  title={Open access database of {EEG} signals recorded during imagined speech},
  author={Pressel Coretto, Germ{\'a}n A and Gareis, Iv{\'a}n E and Rufiner, H Leonardo},
  booktitle={12th International Symposium on Medical Information Processing and Analysis},
  volume={10160},
  pages={1016002},
  year={2017},
  organization={SPIE},
  doi={10.1117/12.2255697}
}

@article{nguyen2018inferring,
  title={Inferring imagined speech using {EEG} signals: a new approach using Riemannian manifold features},
  author={Nguyen, Chuong H and Karavas, George K and Artemiadis, Panagiotis},
  journal={Journal of Neural Engineering},
  volume={15},
  number={1},
  pages={016002},
  year={2018},
  publisher={IOP Publishing},
  doi={10.1088/1741-2552/aa8235}
}

@inproceedings{yang2023biot,
  title={{BIOT}: Cross-data Biosignal Learning in the Wild},
  author={Yang, Chaoqi and Westover, M. Brandon and Sun, Jimeng},
  booktitle={Advances in Neural Information Processing Systems (NeurIPS)},
  volume={36},
  year={2023}
}

@inproceedings{jiang2024labram,
  title={Large Brain Model for Learning Generic Representations with Tremendous {EEG} Data in {BCI}},
  author={Jiang, Wei-Bang and Zhao, Liang-Ming and Lu, Bao-Liang},
  booktitle={The Twelfth International Conference on Learning Representations (ICLR)},
  year={2024}
}

@inproceedings{wang2024eegpt,
  title={{EEGPT}: Pretrained Transformer for Universal and Reliable Representation of {EEG} Signals},
  author={Wang, Guangyu and Liu, Wenchao and He, Yuwei and Xu, Cong and Ma, Lin and Li, Haifeng},
  booktitle={Advances in Neural Information Processing Systems (NeurIPS)},
  volume={37},
  year={2024}
}

@inproceedings{lee2023diffe,
  title={{Diff-E}: Diffusion-based Learning for Decoding Imagined Speech {EEG}},
  author={Lee, Soowon and Lee, Young-Eun and Kim, Heung-Su and Lee, Seong-Whan},
  booktitle={Annual International Conference of the IEEE Engineering in Medicine and Biology Society (EMBC)},
  year={2023},
  doi={10.1109/EMBC40787.2023.10340456}
}

@article{pascualmarqui2002sloreta,
  title={Standardized low-resolution brain electromagnetic tomography ({sLORETA}): technical details},
  author={Pascual-Marqui, Roberto D},
  journal={Methods and Findings in Experimental and Clinical Pharmacology},
  volume={24},
  number={Suppl D},
  pages={5--12},
  year={2002}
}

@article{liu2024msvtnet,
  title={{MSVTNet}: Multi-Scale Vision Transformer Network for {EEG}-Based Motor Imagery Decoding},
  author={Liu, Ke and Yang, Tao and Yu, Zhuliang and Yi, Weibo and Yu, Hao and Wang, Guojun and Wu, Wei},
  journal={IEEE Journal of Biomedical and Health Informatics},
  year={2024},
  doi={10.1109/JBHI.2024.3450753}
}

@article{chen2024eegnex,
  title={Toward reliable signals decoding for electroencephalogram: A benchmark study to {EEGNeX}},
  author={Chen, Xiaorong and Teng, Xiaorong and Chen, Hailing and Pan, Yuyu and Geyer, Philipp},
  journal={Biomedical Signal Processing and Control},
  volume={87},
  pages={105475},
  year={2024},
  publisher={Elsevier},
  doi={10.1016/j.bspc.2023.105475}
}

@article{desikan2006automated,
  title={An automated labeling system for subdividing the human cerebral cortex on {MRI} scans into gyral based regions of interest},
  author={Desikan, Rahul S and S{\'e}gonne, Florent and Fischl, Bruce and Quinn, Brian T and Dickerson, Bradford C and Blacker, Deborah and Buckner, Randy L and Dale, Anders M and Maguire, R Paul and Hyman, Bradley T and Albert, Marilyn S and Killiany, Ronald J},
  journal={NeuroImage},
  volume={31},
  number={3},
  pages={968--980},
  year={2006},
  publisher={Elsevier},
  doi={10.1016/j.neuroimage.2006.01.021}
}

\end{document}